\documentclass{article}

\usepackage{PRIMEarxiv}

\usepackage{times}
\usepackage{natbib}
\usepackage[table]{xcolor}
\usepackage{booktabs}
\usepackage{graphicx}
\usepackage{amsmath,amssymb,amsfonts}
\usepackage{algorithm2e}
\usepackage{cite}
\usepackage{hyperref}
\hypersetup{colorlinks=true,citecolor=teal,linkcolor=blue,urlcolor=blue}
\usepackage{xcolor}
\usepackage{array}
\usepackage{fancyhdr}
\usepackage{lipsum}
\usepackage{rotating}
\usepackage{adjustbox}
\usepackage{siunitx}
\usepackage{float}
\usepackage{multirow}
\usepackage{comment}
\usepackage{textcomp}
\usepackage{placeins}
\usepackage{pifont}
\usepackage{bbm}
\usepackage{mathtools}

\makeatletter
\def\widebreve{\mathpalette\wide@breve}
\def\wide@breve#1#2{%
  \sbox\z@{$#1#2$}%
  \mathop{\vbox{\m@th\ialign{##\crcr
    \kern0.08em\brevefill#1{0.8\wd\z@}\crcr\noalign{\nointerlineskip}%
    $\hss#1#2\hss$\crcr}}}\limits}
\def\brevefill#1#2{$\m@th\sbox\tw@{$#1($}%
  \hss\resizebox{#2}{\wd\tw@}{\rotatebox[origin=c]{90}{\upshape(}}\hss$}

\newcommand*{\bigCup}{\mathop{\mathpalette\big@Cup\relax}\slimits@}
\newcommand*{\big@Cup}[2]{%
   \setbox\z@=\hbox{\m@th$#1\Cup$}%
   \setbox\z@=\vtop{\vbox{\kern.2\ht\z@\copy\z@}\kern.1\ht\z@}%
   \setbox\tw@=\hbox{\m@th$#1\bigcup$}%
   \vcenter{\hbox{\resizebox{!}{1.4\ht\tw@}{\box\z@}}}%
}
\makeatother

\renewcommand{\arraystretch}{1.4}
\title{Meta-Entity Driven Triplet Mining for Aligning Medical Vision-Language Models}

\author{%
  Saban Ozturk\textsuperscript{1,2,3} \And
  Melih B. Yilmaz\textsuperscript{1,2} \And
  Muti Kara\textsuperscript{2,4} \And
  M. Talat Yavuz\textsuperscript{1,2} \And
  Aykut Ko\c{c}\textsuperscript{1,2} \And
  Tolga \c{C}ukur\textsuperscript{1,2,5}\thanks{S. Ozturk and M.B. Yilmaz contributed equally to the study. (Corresponding author: Tolga Çukur, cukur@ee.bilkent.edu.tr)}
}

\date{\today}

\begin{document}
\maketitle

\begin{center}
\textsuperscript{1}Department of Electrical and Electronics Engineering, Bilkent University, Ankara 06800, Turkey\\[0.5ex]
\textsuperscript{2}National Magnetic Resonance Research Center (UMRAM), Bilkent University, Ankara 06800, Turkey\\[0.5ex]
\textsuperscript{3}Department Of Management Information Systems, Ankara Haci Bayram Veli University, Ankara 06570, Turkey\\[0.5ex]
\textsuperscript{4}Department of Computer Engineering, Bilkent University, Ankara 06800, Turkey\\[0.5ex]
\textsuperscript{5}Department of Neuroscience, Bilkent University, Ankara 06800, Turkey
\end{center}

\vspace{2em} 

\begin{abstract}
Diagnostic imaging relies on interpreting both images and radiology reports, but the growing data volumes place significant pressure on medical experts, yielding increased errors and workflow backlogs. Medical vision-language models (med-VLMs) have emerged as a powerful framework to efficiently process multimodal imaging data, particularly in chest X-ray (CXR) evaluations, albeit their performance hinges on how well image and text representations are aligned. Existing alignment methods, predominantly based on contrastive learning, prioritize separation between disease classes over segregation of fine-grained pathology attributes like location, size or severity, leading to suboptimal representations. Here, we propose MedTrim (\underline{M}eta-\underline{e}ntity-\underline{d}riven \underline{Tri}plet \underline{m}ining), a novel method that enhances image-text alignment through multimodal triplet learning synergistically guided by disease class as well as adjectival and directional pathology descriptors. Unlike common alignment methods that separate broad disease classes, MedTrim leverages structured meta-entity information to preserve subtle but clinically significant intra-class variations. For this purpose, we first introduce an ontology-based entity recognition module that extracts pathology-specific meta-entities from CXR reports, as annotations on pathology attributes are rare in public datasets. For refined sample selection in triplet mining, we then introduce a novel score function that captures an aggregate measure of inter-sample similarity based on disease classes and adjectival/directional descriptors. Lastly, we introduce a multimodal triplet alignment objective for explicit within- and cross-modal alignment between samples sharing detailed pathology characteristics. Our demonstrations indicate that MedTrim improves performance in downstream retrieval and classification tasks compared to state-of-the-art alignment methods. Code for MedTrim is publicly available at: \url{https://github.com/icon-lab/MedTrim}.
\end{abstract}


\section{Introduction}
Chest X-ray (CXR) exams play a pivotal role in screening widespread thoracic conditions such as pneumonia, lung cancer, and heart disease. Such clinical assessments rely on the interpretation of both CXR images and radiology reports \citep{10413624}. Yet, the large volume of multimodal CXR data places a heavy burden on physicians, increasing the risk of diagnostic errors, particularly for subtle pathologies that may be small or diffusely distributed across images \citep{limitr}. Recently, medical vision-language models (med-VLMs) have emerged as a powerful approach to efficiently process multimodal imaging data, assisting physicians in clinical decision making \citep{10203795,SHU2024121526,akshay_2}. By integrating visual and textual information, med-VLMs offer the potential for more accurate and comprehensive interpretation of CXR data, ultimately enhancing patient outcomes \citep{9768661, 9638337,amir,Bannur_2023_CVPR}. Despite this promise, however, image-text misalignment in CXR data can cause spurious correlations and noise in multimodal representations, eliciting misleading predictions that can compromise clinical evaluations \citep{sotos,info16020136}. Therefore, alignment methods that offer high-fidelity integration of within- and cross-modal information is critical to sustain high performance and reliability in med-VLMs.

\begin{figure*}
\centering
\includegraphics[width=\textwidth]{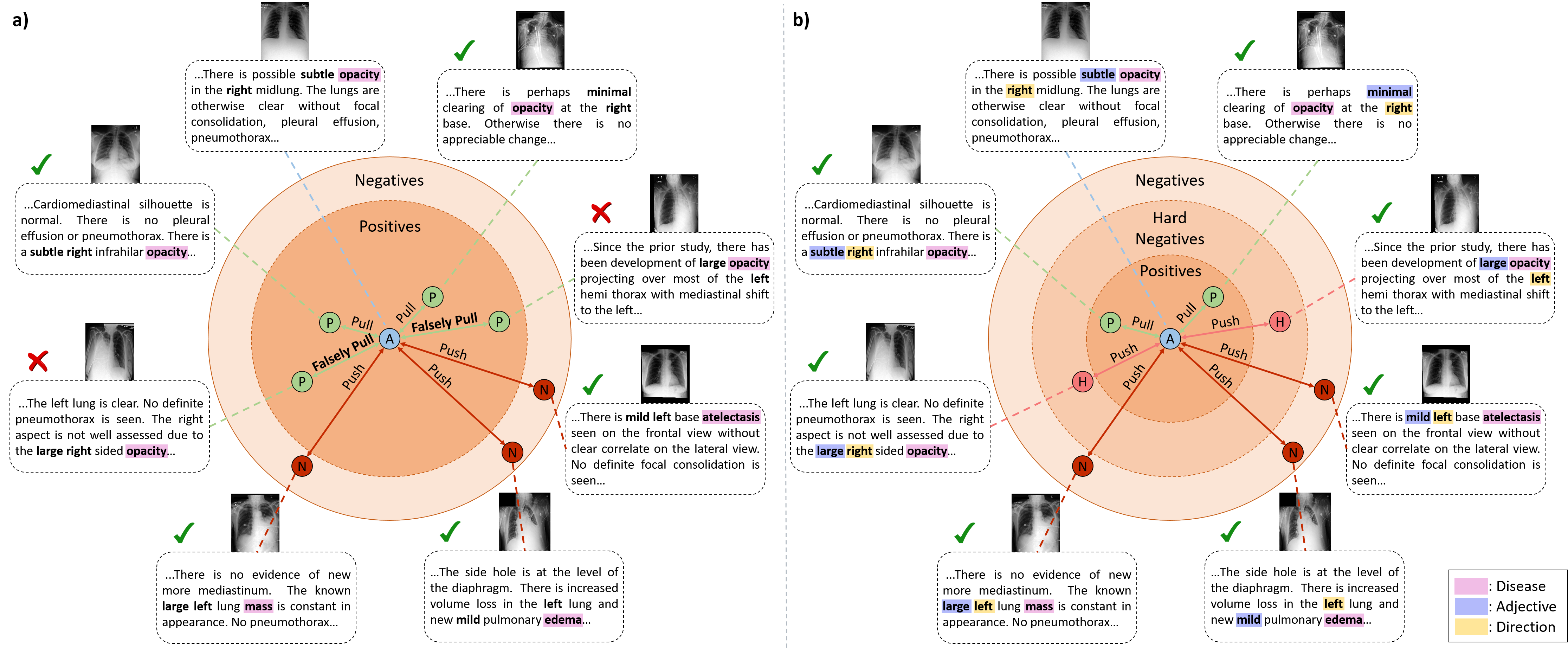}
\caption{Multimodal CXR data carry key attributes pertaining to disease class, and adjectival and directional descriptors of pathology, which can show varying degrees of similarity between data samples. \textbf{a)} Conventional representational learning based on disease classes renders it challenging to dissociate positive samples (with closely aligned attributes) and semi-hard negative samples (with partial alignment of adjectival and directional attributes), forcing undesired alignment between samples with dissimilar attributes. \textbf{b)} To improve alignment, MedTrim leverages a novel multimodal triplet learning framework that explicitly selects positive and semi-hard negative samples, guided by not only disease class but also adjectival and directional descriptors. (A: anchor sample, P: positive sample, H: semi-hard negative sample, N: negative sample, Check mark: accurate alignment, Cross: incorrect alignment.)}
\label{fig:problems}
\end{figure*}

Image-text alignment in med-VLMs has been predominantly conducted via contrastive learning on representations of medical images and radiology reports encoded in a latent embedding space \citep{akshay_1,jlocal2022}. During contrastive learning, random pairs of multimodal data samples (i.e., a sample denoting the CXR image-report for a subject) are selected, and dissimilar samples in each pair are pushed apart—where similarity is controlled via augmentation techniques \citep{10.1007/978-3-031-20059-5_1, CHEN2024103018, 2024arXiv240101583H} or disease labels \citep{2023arXiv230211352V, 2024arXiv240110501L}. A first line of work has proposed to align image and text embeddings at the global and/or local levels via fundamental guidance regarding disease classes. While effective, these methods assess dissimilarity at a relatively coarse level, often relying on learned prototypes \citep{10.1007/978-3-031-19833-5_33} or predefined labels \citep{10.1007/978-3-031-44693-1_35} for diseases. Since contrastive learning optimizes pairwise sample distances to increase inter-class separation, it forces all negative samples (i.e., those from different classes) apart uniformly, treating them as equally dissimilar (Fig. \ref{fig:problems}a). \textit{This can overlook subtle variations in pathology attributes within the same disease class or shared pathological features across different disease classes.}  

An alternative line of work has instead aimed to enhance alignment precision by introducing external knowledge on pathology attributes such as location or size. These methods often extract structured labels from CXR reports in the form of disease-descriptive phrases from medical dictionaries \citep{wang-etal-2023-fine} or prompts from large language models (LLMs) \citep{10.1007/978-3-031-67751-9_7}. A reference semantic similarity measure is then computed by comparing these labels across samples, and contrastive learning is guided by aligning image-text embeddings against the reference. While this approach has been suggested to improve alignment precision compared to using only disease labels, existing methods still rely on the contrastive learning framework. Processing data samples in pairs at each instance, contrastive learning lacks the ability to explicitly select multiple samples with graded degrees of similarity to a given anchor \citep{10203795}. As such, it flattens semantic relations into pairwise sample distances, producing image-text embeddings that carry only implicit semantic information \citep{convirt}. \textit{This can result in over-separation of disease classes at the cost of degradations in intra-class distinctions, rendering suboptimal capture of fine-grained variations in pathology attributes.}

\begin{figure*}
\centering
\includegraphics[width=\textwidth]{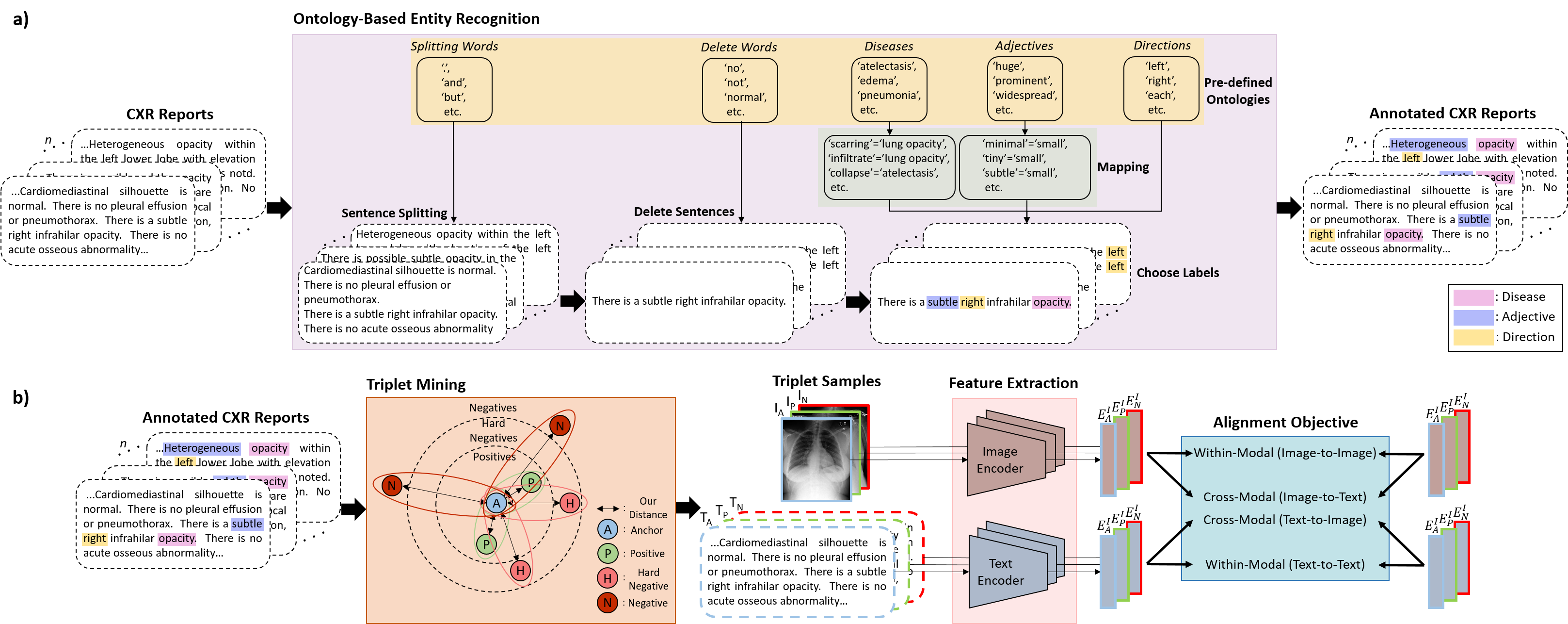}
\caption{\textbf{a)} Given a batch of multimodal CXR samples, MedTrim first deploys its OBER module to identify disease labels along with adjectival and directional descriptors of corresponding pathology from radiology reports. \textbf{b)} Using these pathology attributes as meta entities, MedTrim performs triplet mining to select positive and semi-hard negative samples via entity-based weighting with respect to the anchor sample. The embeddings of the selected triplets are extracted with transformer-based image and text encoders. For image-text alignment, the encoders are fine-tuned via a multimodal triplet alignment objective that synergizes within-modal and cross-modal alignment expressed over embedding vectors.}
\label{fig:main_MedTrim}
\end{figure*}

In this study, we propose a novel med-VLM alignment technique, MedTrim, that leverages meta-entity driven triplet mining to enable explicit guidance from key pathology attributes. Since CXR datasets generally lack detailed pathology annotations, MedTrim first introduces an ontology-based entity recognition (OBER) module to extract pathology-specific meta-entities pertaining to disease class, adjectival descriptors (e.g., `mild', `severe') and directional descriptors (e.g., `upper lobe', `peripheral') from radiology reports (Fig. \ref{fig:main_MedTrim}). To address the limitations of ubiquitous contrastive methods, MedTrim then employs a new multimodal framework based on triplet learning\footnote{\textbf{Triplet learning:} A representational learning approach that explicitly describes relationships between data triplets comprising anchor, positive and negative samples.} that mines positive samples with heavily-matched meta-entities to the anchor sample, and semi-hard negative samples with moderate differences in detailed pathology attributes (Fig. \ref{fig:problems}b). To this end, MedTrim introduces a novel score function that assesses the aggregate inter-sample similarity based on the examined meta-entities. Driven by this score function, explicit sample selection promotes active learning of subtle pathology variations in CXR data, while avoiding over-clustering of samples within individual disease classes. Unlike many previous alignment objectives constrained to cross-modal terms, MedTrim expresses a unified triplet alignment objective balancing preservations of within-modal (i.e., image-to-image, text-to-text) and cross-modal (i.e., image-to-text, text-to-image) relationships. Through comprehensive experiments on public CXR datasets, we demonstrate that these technical advances enable MedTrim to improve precision over state-of-the-art alignment methods, in turn boosting performance in downstream multimodal retrieval and disease classification tasks.

\subsubsection*{\textbf{Contributions}}
\begin{itemize}
    \item To our knowledge, MedTrim is the first alignment technique for med-VLMs that employs triplet mining based on meta-entities of multimodal CXR data.
    \item MedTrim introduces a new ontology-based entity recognition module to faithfully extract disease labels, directional and adjectival descriptors from radiology reports.
    \item MedTrim introduces a novel score function for entity-guided triplet mining that captures an aggregate measure of inter-sample similarity.
    \item MedTrim introduces a multimodal triplet alignment objective tailored to optimize image and text representations for effective within- and cross-modal alignment. 
\end{itemize}

\section{Related Work}
\textbf{Medical vision-language models.} In recent years, med-VLMs have gained prominence in CXR evaluation by integrating medical images with radiology reports \citep{pmlr-v139-radford21a, pmlr-v139-jia21b}. For joint analysis of multimodal data, these models encode image and text samples into a shared latent embedding space. Early approaches employed convolutional neural networks as image encoders and recurrent neural networks as text encoders \citep{7298932,faghri2017vse++}. Yet, transformer-based encoders have since become the standard for both modalities due to their superior representational capacity \citep{2024arXiv240112208C, 10.1145/3583780.3614961, 2021arXiv210316022W, 2022arXiv221006044W, 2023arXiv230519894W, nezhad2025generativeautoregressivetransformersmodelagnostic, atli2024i2imambamultimodalmedicalimage}. Despite their powerful encoding capabilities, med-VLMs often struggle with naive combinations of image and text embeddings, leading to suboptimal performance. CXR images capture the full range of normal anatomical structures and pathological abnormalities in a given subject, whereas radiology reports often provide condensed, context-dependent descriptions that emphasize diagnostically critical abnormalities \citep{lee2025cxr}. Moreover, substantial variability in radiological prose introduces additional discrepancies \citep{tanno2024collaboration}. These inherent image-text misalignments can cause med-VLM-based evaluations to be inaccurate or clinically irrelevant \citep{10.1007/978-3-031-16443-9_69, PARK2024103021, 10376864, zhang2023knowledge, 10182304}. Therefore, ensuring proper alignment between image and text representations is critical for training reliable med-VLMs.

\vspace{0.8ex}
\textbf{Conventional contrastive learning.} Coming forth as the mainstream alignment framework, contrastive learning that performs pairwise sample processing has been extensively adopted for med-VLMs \citep{convirt,9710099,2022arXiv221006044W}. Given an anchor sample from a batch, positive samples from the same disease class are pulled closer to the anchor, while negative samples from different classes are pushed away \citep{YANG202271,2023arXiv231007027L,2024arXiv240101583H}. Information regarding disease classes can be defined through prototype clustering of multimodal CXR data or predefined labels \citep{2023arXiv230211352V,2024arXiv240110501L}; and contrastive objectives can be combined with masked-image or masked-text modeling to further enhance learning efficacy \citep{10.1007/978-3-031-20059-5_1,2024arXiv240101591L,CHEN2024103018}. While some methods align image-text embeddings at a global level, others have also sought local alignment between image patches and text fragments, aggregating information across candidate patches/fragments through pooling or fusion mechanisms \citep{ZHANG2023107522,10376847}. Despite its ubiquity, conventional contrastive learning primarily segregates disease classes broadly, making the aligned image-text embeddings relatively insensitive to fine-grained distinctions among pathology attributes.

\vspace{0.8ex} \textbf{Knowledge-guided contrastive learning.} To improve alignment precision in med-VLMs, knowledge-guided contrastive learning methods have been proposed in recent studies. These methods commonly represent knowledge of pathology attributes using structured labels extracted from CXR reports \citep{10445007}. Knowledge extraction has been performed through various approaches, including disease descriptions from medical dictionaries and learned entity extractors, as well as general-purpose named-entity recognition tools and language models \citep{2022arXiv221010163W,zhang2023knowledge,2024arXiv240309294L}. Structured labels are then compared across samples to derive a reference measure of semantic similarity, and image-text embeddings are aligned against this reference \citep{10376864}. Although attribute-based sample relationships can implicitly influence the learning objective in this approach, contrastive learning lacks the ability to explicitly select samples reflecting fine-grained variations with respect to the anchor, as it performs pairwise processing that inevitably flattens semantic relationships into pairwise distances. As a result, gradual variations in pathology attributes are rendered implicit and thereby difficult to capture \citep{yin-etal-2025-kia}. In addition, many previous knowledge-guided methods have exclusively used cross-modal objectives expressed by comparing image against text embeddings. Neglecting within-modal objectives can cause suboptimal image-image or text-text alignment for similar CXR samples, and compromise performance in unimodal tasks (e.g., image-to-image retrieval) \citep{OZTURK2023118938}. 
\vspace{0.8ex} 

\textbf{Proposed technique.} MedTrim is a novel multimodal triplet learning framework that enables explicit guidance from key pathology attributes extracted from CXR reports. A recent alignment method based on contrastive learning has augmented cross-entropy loss with a ranking loss, with positive samples taken as image/text pairs from the same subject and negative samples taken as image/text from different subjects \citep{second_met}. While the ranking loss simultaneously specifies positive and negative samples akin to a triplet loss, this previous method does not use information on disease classes or pathology attributes for guided sample selection. Differing from existing alignment methods, MedTrim leverages explicit guidance from pathology attributes in selection of positive and semi-hard negative samples for each anchor. As such, it improves capture of fine-grained representations of intra-class variations, while preventing over-clustering in the latent embedding space. Unlike many knowledge-guided alignment methods, it formulates a unified objective that incorporates both within-modal and cross-modal triplet alignment terms for a faithful representation of both unimodal and multimodal relationships in CXR data.

\section{Theory}
\subsection{Overview of MedTrim}
MedTrim is devised to align image-text representations for multimodal CXR data. Let $\mathbbm{D}=\left\{ \left( I,T \right)_{1},...,\left( I,T \right)_{K} \right\}$ denote a training dataset with CXR images and reports from $K$ subjects, where $I\in \mathbb{R}^{H \times W}$ are single-view images with $H$ and $W$ specifying height and width, and $T$ are reports comprising $J$ tokens $\left\{ t_{1},t_{2},...,t_{J} \right\}$ (where tokens are single words and $J$ is variable across reports). For each training iteration, MedTrim draws a distinct mini-batch with $k$ image-text samples from $\mathbbm{D}$ denoted as $\mathbbm{B}=\left\{ \left( I,T \right)_{1},...,\left( I,T \right)_{k} \right\}, k\ll K$. The four training stages per mini-batch are given below: 
\vspace{0.2ex}

$\triangleright$ In the first stage, MedTrim leverages its OBER module (Sec. \ref{sec:OBER}) to extract a set of meta-entities from each radiology report as $\mathbbm{m}=\mathrm{OBER}\left( T \right)$. The meta-entities for each report are cast as a structured list $\mathbbm{m}=\left( \mathbf{d},adj(\mathbf{d}),dir(\mathbf{d}) \right)$, where $\mathbf{d}$ is a list of diseases classes, $adj(\cdot)$ is a disease-specific list of adjectival descriptors, and $dir(\cdot)$ is a disease-specific list of directional descriptors. Note that each report can contain multiple disease classes, and there can be multiple adjectival or directional descriptors for each disease class.   

$\triangleright$ In the second stage, MedTrim performs entity-driven triplet mining within the given mini-batch (Sec. \ref{sec:Miner}). A total of $n_{tri}$ triplets are formed, where the $n$th triplet is composed as $\mathbbm{Tri}_n=\{(I_A,T_A),(I_P,T_P),(I_N,T_N)\}_n$ with $A$ denoting the anchor sample, $P$ denoting the positive sample, and $N$ denoting the negative sample. MedTrim leverages an entity similarity score, $score$, used to mine semi-hard negative samples for improved sensitivity to fine-grained variations in CXR data.  
\vspace{0.5ex} 

$\triangleright$ In the third stage, the mined triplets are encoded onto a latent embedding space via transformer-based image and text encoders (Sec. \ref{sec:Encoder}). This results in an embedding vector per image $E^I = \mathrm{Trans}(I) \in \mathbb{R}^{c}$ that globally pools information across image patches, and an embedding vector per report $E^T = \mathrm{Trans}(T) \in \mathbb{R}^{c}$ that pools information across tokens ($c$: embedding dimensionality). As such, each mined triplet is projected onto image representations $\left\{ E_{A}^{I},E_{P}^{I},E_{N}^{I} \right\}$ and text representations $\left\{ E_{A}^{T},E_{P}^{T},E_{N}^{T} \right\}$.
\vspace{0.5ex}

$\triangleright$ In the fourth stage, MedTrim expresses a unified triplet objective $\mathcal{L}_{MedTrim}$ to align image-text representations (Sec. \ref{sec:Objective}). $\mathcal{L}_{MedTrim}$ contains separate triplet terms for within-modal alignment (image-to-image, text-to-text) and cross-modal alignment (image-to-text, text-to-image). This unified objective enables MedTrim to precisely align CXR images against respective radiology reports, without compromising representations of within-modal relationships.

\SetKwInput{KwData}{Input}
\SetKwInput{KwResult}{Output}
\SetKwInput{KwDo}{Do}
\RestyleAlgo{ruled}
\SetAlgoNoLine
\SetKwInOut{Parameters}{Parameters}
\begin{algorithm}[t]
    \small
    \caption{OBER Module}
    \label{alg:one}
    \KwData{$I$: CXR image, $T$: radiology report\\
    \Indp\Indp $O_{dis}$, $O_{adj}$, $O_{dir}$, $O_{spl}$, $O_{del}$: disease, adjective, direction, sentence splitting, and delete words.\\    
    $f{(\cdot)}$: denotes various mappings for word extraction}
    \KwResult{$\mathbbm{m}=\left( \mathbf{d},adj(\mathbf{d}),dir(\mathbf{d}) \right)$} 
    \KwDo{\\\vspace{0.05cm}
            Lemmatize report$\,\rightarrow\,$ $\bar{T}=f_{lem}(T)$ \\ \vspace{0.05cm}
            Split into sentences$\,\rightarrow\,$$\{ s_1,..., s_V \}=f_{spl}
            (\bar{T},O_{spl})$ \\ \vspace{0.05cm}
            Delete irrelevant$\,\rightarrow\,$$\{ s_1,..., s_v \}=f_{del}
            (\{ s_1,..., s_V \},O_{del})$\\ \vspace{0.05cm}
            \For{$i=\text{}{1}$:$v$}{
                Disease class for sentence $i$$\,\rightarrow\,$$\mathbf{d}[i]$=$f_{dis}(s_i,O_{dis})$\\ \vspace{0.05cm}
                \textbf{if} $\mathbf{d}[i]$ exists: \\ \vspace{0.05cm}
                    \Indp Adjectival entity list$\,\rightarrow\,$$adj(\mathbf{d}[i])$=$f_{adj}(s_i,O_{adj})$\\ \vspace{0.05cm}
                    Directional entity list$\,\rightarrow\,$$dir(\mathbf{d}[i])$=$f_{dir}(s_i,O_{dir})$\\
                }
     }
     \Return $\left( \mathbf{d}, adj(\mathbf{d}), dir(\mathbf{d}) \right)$
\end{algorithm}
\subsection{Ontology-Based Entity Recognition}
\label{sec:OBER}
Public CXR databases primarily provide raw radiology reports and disease labels, but they often lack isolated labels describing pathology attributes. Learning-based methods for attribute extraction are typically prone to unwanted hallucinations that can compromise accuracy \citep{2024arXiv240610185C}. Meanwhile, many ontology-based extractors are either built for general language processing \citep{10.1371/journal.pone.0116656}, potentially compromising sensitivity on medical data, or they are not publicly accessible. To address this gap, we introduce a new ontology-based recognition module (OBER) for high-fidelity extraction of meta-entities from CXR reports as outlined in Alg. \ref{alg:one}. We devised OBER as an ontology with disease words ($O_{dis}$), adjective words ($O_{adj}$), and direction words ($O_{dir}$) to detect meta-entities, as well as sentence splitting words ($O_{spl}$) to control text fragmentation, and delete words ($O_{del}$) to discard irrelevant text fragments. These words were curated from a medical dictionary, manually extended to include a comprehensive list of synonyms \citep{irvin2019chexpert}, and used in lemmatized form to eliminate the effects of various inflections such as singular-plural distinctions \citep{toporkov2024role}. The resultant ontology comprised synsets for 12 disease, 98 adjective, 4 direction, 6 splitting, and 16 delete words. 

Given a radiology report $T$, OBER first lemmatizes it by processing its individual tokens as $\bar{T}=f_{lem}(T)$. It then splits the report into sentences using splitting words as $\{s_1,...,s_{V}\}=f_{spl}(\bar{T},O_{spl})$, where $V$ is the number of sentences in $\bar{T}$. Sentences that contain irrelevant text are discarded using delete words to obtain a condensed list of $v$ sentences: $\{s_1,...,s_{v}\}=f_{del}(\{s_1,...,s_{V}\},O_{del})$. Following this clean-up, remaining sentences are processed to extract meta-entities starting from empty sets, $\mathbf{d}=\varnothing,\, adj(\mathbf{d})=\varnothing,\, dir(\mathbf{d})=\varnothing$. For the $i$th sentence, a search for disease words is conducted to recognize the disease class: $\mathbf{d}[i] = \{f_{dis}(s_i,O_{dis})\}$, where $f_{dis}$ returns an empty set when no disease words are present in the sentence. If a disease class is recognized, then further search for adjective and direction words is conducted as $adj(\mathbf{d}[i]) = f_{adj}(s_i,O_{adj})$ and $dir(\mathbf{d}[i]) = f_{dir}(s_i,O_{dir})$. Note that adjectival and directional searches can return empty sets, single words or multiple words describing the pathology for the given disease class $\mathbf{d}[i]$. Processing of all sentences produces a list for disease classes $\mathbf{d}=\{\mathbf{d}[1],\mathbf{d}[2],...\}$, a list for adjectival descriptors $adj(\mathbf{d})= \{adj(\mathbf{d}[1]),adj(\mathbf{d}[2]),...\}$, and a list for directional descriptors $dir(\mathbf{d}) = \{dir(\mathbf{d}[1]),dir(\mathbf{d}[2]),...\}$. These lists are aggregated to form a structured list of meta-entities for each report, $\mathbbm{m}=\left( \mathbf{d},adj(\mathbf{d}),dir(\mathbf{d}) \right)$.

\subsection{Entity-Driven Triplet Mining}
\label{sec:Miner}
\textbf{Score function:} MedTrim leverages a novel entity-driven triplet mining procedure to select samples that accurately capture fine-grained variations within individual disease classes. In this procedure, a random image-text sample from a given mini-batch is taken as the anchor sample. Afterwards, positive and negative samples are selected from the mini-batch to create a triplet based on similarity of meta-entities. Note that each sample can carry labels for multiple diseases, and there may be multiple adjectival and directional descriptors within each disease class. These hierarchically-structured entities can show varying levels of match between CXR samples, and evaluating similarity of meta-entities hence requires a carefully constructed measure that respects the entity hierarchy. To address this issue, we introduce a novel score function that measures the aggregate similarity of disease classes, adjectival descriptors and directional descriptors between two CXR samples as follows:
\begin{align}
score(\mathbbm{m}_i,\mathbbm{m}_j) = &\frac{\delta_{\mathbf{d}}(\mathbf{d}_i \cap \mathbf{d}_j)}{|\mathbf{d}_i \cup \mathbf{d}_j|} \times \sum_{\mathbf{d}[q] \,\in\, \mathbf{d}_i \cap \mathbf{d}_j} \Big( \frac{\gamma_{0} + \gamma_{1}\mathrm{JI}_{adj}(\mathbf{d}[q]) + \gamma_{2}\mathrm{JI}_{dir}(\mathbf{d}[q])}{\gamma_{0} + \gamma_{1}\delta_{adj}(\mathbf{d}[q]) + \gamma_{2}\delta_{dir}(\mathbf{d}[q])}\Big), \label{eq:score}
\end{align}

where $\mathbbm{m}_{i}$, $\mathbbm{m}_{j}$ are meta-entities and $\mathbf{d}_{i}$, $\mathbf{d}_{y}$ are the label sets of disease classes for the $i$th and $j$th samples, $\cup$ is the union operator that pools labels across sets, and $\cap$ is the intersection operator that returns the common labels across sets. 

The proposed score function identifies disease class labels shared between two CXR samples, and in case of a matching disease class, it proceeds onto evaluating potential matches between the labels for adjectival and directional descriptors. Accordingly, the summation in Eq. \ref{eq:score} is taken over disease classes, where $\mathbf{d}[q]$ denotes the $q$th disease class in the intersection set of $\mathbf{d}_i$ and $\mathbf{d}_j$. For a given $\mathbf{d}[q]$ shared between $i$th and $j$th samples, the score is a combination of the degree of label matches for disease class (taken as unity), adjectival descriptors and directional descriptors, with $\gamma_{0,1,2}$ denoting the respective weighting parameters such that $\gamma_{0}+\gamma_1+\gamma_2=1$. Meanwhile, $\mathrm{JI}_{adj,dir}$ denote Jaccard indices that reflect the label matches for adjectival and directional descriptors \citep{bertels2019optimizing}: 
\begin{align}
\mathrm{JI}_{adj}(\mathbf{d}[q]) = \frac{|adj_i(\mathbf{d}[q]) \cap adj_j(\mathbf{d}[q])|}{|adj_i(\mathbf{d}[q]) \cup adj_j(\mathbf{d}[q])|}, \\
\mathrm{JI}_{dir}(\mathbf{d}[q]) = \frac{|dir_i(\mathbf{d}[q]) \cap dir_j(\mathbf{d}[q])|}{|dir_i(\mathbf{d}[q]) \cup dir_j(\mathbf{d}[q])|}.
\end{align}
Note that Jaccard indices have a range of $0\le \mathrm{JI} \le1$. To ensure that the weighted score for $\mathbf{d}[q]$ is also normalized to the same range, the summation denominator in Eq. \ref{eq:score} contains $\delta_{adj}$ to indicate whether the union set of adjectival descriptors for the disease class $\mathbf{d}[q]$ is non-empty, and $\delta_{dir}$ to indicate whether the union set of directional descriptors for the disease class ${dir}[q]$ is non-empty. An indicator function for disease class, $\delta_{\mathbf{d}}$, is also included to assign a zero score to cases when disease class labels do not show any overlap:  
\begin{align}
    &\delta_{\mathbf{d}}(\mathbf{d}_i \cap \mathbf{d}_j) = \mathrm{sgn}(|\mathbf{d}_i \cap \mathbf{d}_j|), \\
    &\delta_{adj}(\mathbf{d}[q]) = \mathrm{sgn}(|adj_i(\mathbf{d}[q]) \cap adj_j(\mathbf{d}[q])|),\\
    &\delta_{dir}(\mathbf{d}[q]) = \mathrm{sgn}(|dir_i(\mathbf{d}[q]) \cap dir_j(\mathbf{d}[q])|),
\end{align}
where $\mathrm{sgn}$ is the signum operator.

\vspace{0.8ex} \textbf{Sample selection:} Given a mini-batch $\mathbbm{B}$ with $k$ samples of CXR data, MedTrim mines a total of $n_{tri}$ triplets of the form $\mathbbm{Tri}_n=\{(I_A,T_A), (I_P,T_P), (I_N,T_N)\}_n$ ($n$: triplet index). Each triplet is devised to contain a unique anchor sample $A$, randomly drawn from $\mathbbm{B}$. Assume that the $n$th triplet is initiated by drawing the $q$th mini-batch sample as $A$, such that $(I_A,T_A)=\mathbbm{B}[q]=(I,T)_q$ and $\mathbbm{m}_A=\mathbbm{m}_q$. MedTrim employs Eq. \ref{eq:score} to compute the similarity scores between $A$ and remaining samples within the mini-batch. The image-text for the positive sample $P$ are drawn from $\mathbbm{B}$ by identifying the sample with maximum similarity score to $A$:
\begin{align}
(I_P,T_P) = \mathbbm{B}\Bigg[ \underset{j\in [1 \mbox{ } k],\,\,j\neq q}{\text{argmax}} \left| score(\mathbbm{m}_A,\mathbbm{m}_j) \right| \Bigg]. 
\end{align}
A crucial factor that influences the ability of triplet mining to learn discriminative representations is the selection of negative samples. While easy negatives that are substantially dissimilar from the anchor present trivial cases for representational learning, hard negatives that closely resemble positive samples can elicit underfitting when distinctions become excessively subtle. To improve learning of intra-class variations, MedTrim instead employs semi-hard negative samples that present challenging cases while still bearing moderate similarity to positive samples. Accordingly, the image-text for the semi-hard negative sample $N$ are drawn from $\mathbbm{B}$ by identifying the sample with minimum score in the range $[\tau_{min} \,\mbox{ }\, \tau_{max}]$:
\begin{align}
(I_N,T_N) &= \mathbbm{B} \Bigg[ \underset{j\in [1 \mbox{ } k],\,\,j\neq q}{\text{argmin}} \left| score(\mathbbm{m}_A,\mathbbm{m}_j) \right| \nonumber \quad \mbox{ s.t. } \tau_{min} \leq score \leq \tau_{max} \Bigg].
\end{align}
The range constraint ensures that $N$ shows a degree of match to $A$ in terms of disease classes, albeit that has mismatched adjectival and/or directional descriptors. This semi-hard negative selection enforces the model to better capture fine-grained intra-class variations due to varying pathology attributes. 

\subsection{Encoding onto Latent Embedding Space}
\label{sec:Encoder}
Following entity-driven triplet mining, multimodal CXR data are projected onto a latent embedding space, where the alignment objective is expressed. MedTrim leverages transformer-based encoders for both CXR images and reports, given the high representational power to the self-attention mechanism. To construct the encoder inputs, a given CXR image $I$ is split into $p$ patches to form the sequence $e^{I}=\{e^I_1,e^I_2,...,e^I_p\}$, and a given CXR report $T$ is split into $p$ tokens (i.e., words) to form the sequence $e^{T}=\{e^T_1,e^T_2,...,e^T_p\}$. For brevity, here we will present a common description of the encoder processing for the two modalities. Accordingly, the input sequence $e$ is first embedded via a learnable linear layer (Lin) and supplemented with learnable position encodings (PE) \citep{dosovitskiy2020image}:
\begin{align}
h^{(0)} = \text{Lin}(e) + \text{PE}(e).
\end{align}
The embedded sequence is then projected through $L$ transformer blocks, comprising normalization (LN), multi-head self-attention (MHSA), and multi-layer perceptron (MLP) layers along with residual connections  \citep{dosovitskiy2020image}. The mappings across the $l$th block can be expressed as:
\begin{align}
\tilde{h}^{(l)} &= \text{MHSA}(\text{LN}(h^{(l-1)}))+h^{(l-1)},\\
h^{(l)} &= \text{MLP}(\text{LN}(\tilde{h}^{(l)}))+\tilde{h}^{(l)},
\end{align}
where $h^{(l)}$ is the block output. At the output of the $L$th (i.e., final) block, the hidden representations for sequence elements are pooled \citep{dosovitskiy2020image}: 
\begin{align}
E = \text{GAP}(h^{(L)}),
\end{align}
where $\text{GAP}$ denotes global average pooling across sequence elements. The overall mapping that encoders perform can be summarized as: $E^I=\mathrm{Trans}(I)$ and $E^T=\mathrm{Trans}(T)$.

\subsection{Multimodal Triplet Alignment Objective}
\label{sec:Objective}
Conventional triplet learning performs unimodal representational learning by pulling together similar samples (i.e., anchor and positive), while pushing away dissimilar samples (i.e., anchor and negative) in the latent embedding space. For unimodal embedding vectors $\{E_A,E_P,E_N\}$, the standard triplet learning objective is given as \citep{10467142}:
\begin{equation}
f_{tri}\left(E_A, E_P, E_N \right)= \text{max}\left( 0,\mathrm{cos} \left( E_A, E_P \right) - \right.\left. \mathrm{cos} \left( E_A, E_N \right) + \alpha \right),
\label{eq:triplet-loss}
\end{equation}
where $\mathrm{cos}(\cdot)$ denotes the cosine similarity between embedding vectors, and $\alpha$ is a margin parameter. Adopting representational learning frameworks for image-text alignment, many previous alignment methods in the literature have expressed cross-modal learning objectives \citep{YANG202271}. Thus, a straightforward generalization of triplet learning to align multimodal CXR data can be implemented via the following terms: 
\begin{align}
\mathcal{L}_{I2T} = f_{tri}\left(E^I_A, E^T_P, E^T_N \right), \label{eq:triplet-cross1}\\
\mathcal{L}_{T2I} = f_{tri}\left(E^T_A, E^I_P, E^I_N \right), \label{eq:triplet-cross2}
\end{align}
where $I2T$ denotes image-to-text and $T2I$ denotes text-to-image alignment. Note, however, that solely using the cross-modal terms will attempt to align image against text embeddings, while neglecting within-modal alignment among images or among text. To seek image-text alignment without compromising an accurate capture of modality-specific relationships in CXR data, MedTrim leverages a new multimodal triplet learning objective that also includes within-modal terms:
\begin{align}
\mathcal{L}_{I2I} = f_{tri}\left(E^I_A, E^I_P, E^I_N \right), \label{eq:triplet-within1}\\
\mathcal{L}_{T2T} = f_{tri}\left(E^T_A, E^T_P, E^T_N \right), \label{eq:triplet-within2}
\end{align}
where $I2I$ denotes image-to-image, and $T2T$ denotes text-to-text alignment. Finally, the overall objective for MedTrim can then be expressed as follows: 
\begin{align}
\mathcal{L}_{MedTrim} = \eta \left( \mathcal{L}_{I2T} + \mathcal{L}_{T2I} \right) +(1-\eta) \left( \mathcal{L}_{I2I} + \mathcal{L}_{T2T} \right),
\end{align}
where \(\eta\) is a weighting parameter that balances the contributions of cross-modal versus within-modal alignment terms.

\begin{table}[t]
\centering
\caption{\textbf{Ablation study on the meta-entity guidance.} 
MedTrim was compared against variants that ablated guidance 
from both adjectival and directional descriptors, from adjectival descriptors, 
and from directional descriptors. Consistency of retrieved items was examined in terms of disease-class (Dis.), 
adjectival-descriptor (Adj.), and directional-descriptor (Dir.). Average performances across I2I, I2T, T2I, T2T tasks 
are reported as P@R (\%) where R is the number of retrieved samples. Boldface marks the top-performing method.}
\resizebox{0.6\columnwidth}{!}{%
\begin{tabular}{ccc ccc ccc ccc} 
\toprule
\multicolumn{3}{c}{\textbf{Meta Entity}} 
& \multicolumn{3}{c}{\textbf{P@10}} 
& \multicolumn{3}{c}{\textbf{P@20}} 
& \multicolumn{3}{c}{\textbf{P@50}} \\
\cmidrule(lr){1-3}\cmidrule(lr){4-6}\cmidrule(lr){7-9}\cmidrule(lr){10-12}
\textbf{Dis.} & \textbf{Adj.} & \textbf{Dir.}
& \textbf{Dis.} & \textbf{Adj.} & \textbf{Dir.}
& \textbf{Dis.} & \textbf{Adj.} & \textbf{Dir.}
& \textbf{Dis.} & \textbf{Adj.} & \textbf{Dir.} \\
\midrule
\checkmark&&                      & 95.6 & 81.5 & 85.3 & 96.4 & 82.6 & 85.4 & \textbf{97.3} & 83.5 & 88.3 \\
\checkmark&\checkmark&            & 95.7 & 84.2 & 84.8 & 95.8 & 86.1 & 84.9 & 96.9 & 86.7 & 88.5 \\
\checkmark&&\checkmark            & 95.4 & 81.3 & 88.5 & 95.7 & 83.5 & 88.9 & 96.8 & 83.4 & 91.6 \\
\checkmark&\checkmark &\checkmark & \textbf{95.9} & \textbf{84.5} & \textbf{89.3} & \textbf{96.4} & \textbf{86.0} & \textbf{90.0} & 97.2 & \textbf{87.8} & \textbf{92.2} \\
\bottomrule
\end{tabular}
}
\label{table:metaentity}
\end{table}

\section{Experimental Methods}
\subsection{Datasets}
In our experiments, we used three public datasets: MIMIC-CXR \citep{johnson2019mimic}, CheXpert \citep{irvin2019chexpert}, and RSNA Pneumonia \citep{doi:10.1148/ryai.2019180041}. All alignment methods were trained on MIMIC-CXR. Afterwards, they were evaluated for retrieval tasks on MIMIC-CXR, and zero-shot disease classification tasks on CheXpert and RSNA. In MIMIC-CXR, 50,000 paired samples of CXR images and radiology reports were analyzed \citep{johnson2019mimic}, covering 12 diseases (Atelectasis, Cardiomegaly, Consolidation, Edema, Enlarged Card., Fracture, Lung Lesion, Lung Opacity, Pleural Effusion, Pleural Other, Pneumonia, Pneumothorax). Data were split randomly into training (70\%), validation (10\%), test (20\%) sets without any subject overlap. Models were learned on the training set, while hyperparameter selection was conducted on the validation set. Retrieval tasks were executed on 14,000 samples in the test set of MIMIC-CXR. In CheXpert, a subset CheXpert 5x200 was constructed following the procedures in \citep{convirt}, which contained 200 images from each of five disease classes: Atelectasis, Cardiomegaly, Edema, and Pleural Effusion. In RSNA Pneumonia, 4,000 CXR image samples drawn from NIH's CXR database were analyzed \citep{doi:10.1148/ryai.2019180041}, while preserving the original label distribution for pneumonia (20\%) and healthy controls (80\%). Note that all samples in CheXpert and RSNA were used as the test set for evaluating zero-shot disease classification.

\begin{table*}[t]
\centering
\caption{\textbf{Ablation study on the multi-modal triplet alignment objective.} Variants were formed by selectively including the following components in the objective: triplet learning ($f_{tri}$, \ding{55}: contrastive, \checkmark: triplet), within-modal alignment losses ($\mathcal{L}_{I2I+T2T}$), image-to-text alignment loss ($\mathcal{L}_{I2T}$), and text-to-image alignment loss ($\mathcal{L}_{T2I}$). Average retrieval performances across the three meta-entities are reported as P@R.}
\resizebox{1.0\columnwidth}{!}{%
\begin{tabular}{c c c c c c c c c c c c c c c c}
\toprule
\multicolumn{4}{c}{\textbf{Loss components}} & \multicolumn{3}{c}{\textbf{I2I}} & \multicolumn{3}{c}{\textbf{I2T}} & \multicolumn{3}{c}{\textbf{T2I}} & \multicolumn{3}{c}{\textbf{T2T}}\\
\cmidrule(lr){1-4} \cmidrule(lr){5-7} \cmidrule(lr){8-10} \cmidrule(lr){11-13} \cmidrule(lr){14-16}
$f_{tri}$ & $\mathcal{L}_{I2I+T2T}$ & $\mathcal{L}_{I2T}$ & $\mathcal{L}_{T2I}$ & \textbf{P@10} & \textbf{P@20} & \textbf{P@50} & \textbf{P@10} & \textbf{P@20} & \textbf{P@50} & \textbf{P@10} & \textbf{P@20} & \textbf{P@50} & \textbf{P@10} & \textbf{P@20} & \textbf{P@50} \\ 
\midrule
\ding{55} & \checkmark &  &  & 81.6 & 82.1 & 82.4 & 73.7 & 74.3 & 75.1 & 76.6 & 77.0 & 77.1 & 82.0 & 82.3 & 82.6 \\
\ding{55} & \checkmark & \checkmark &  & 88.6 & 89.1 & 89.5 & 84.8 & 85.0 & 85.3 & 83.6 & 84.2 & 84.8 & 89.3 & 89.8 & 90.1 \\
\ding{55} & \checkmark &  & \checkmark & 87.5 & 88.1 & 88.7 & 82.3 & 82.8 & 83.2 & 87.2 & 87.8 & 88.5 & 90.0 & 90.5 & 91.0 \\
\ding{55} &  & \checkmark & \checkmark & 86.6 & 87.0 & 87.2 & 85.5 & 86.0 & 86.3 & 87.0 & 87.5 & 88.2 & 88.3 & 89.0 & 89.6 \\
\ding{55} & \checkmark & \checkmark & \checkmark & 90.7 & 90.8 & 91.2 & 89.9 & 90.4 & 90.7 & 90.1 & 90.5 & 91.0 & 93.2 & 94.5 & 95.1 \\
\midrule
\checkmark & \checkmark & & & 86.6 & 87.1 & 87.4 & 80.7 & 81.3 & 81.4 & 82.6 & 83.3 & 83.9 & 90.3 & 91.1 & 92.6 \\
\checkmark & \checkmark & \checkmark & & 94.7 & 95.0 & 95.5 & 93.1 & 93.5 & 93.8 & 91.8 & 92.0 & 92.1 & 93.1 & 94.1 & 94.8\\
\checkmark & \checkmark & & \checkmark & 94.9 & 95.1 & 95.4 & 92.5 & 92.9 & 93.2 & 93.0 & 93.4 & 93.5 & 93.6 & 94.4 & 95.1\\
\checkmark &  & \checkmark & \checkmark & 93.6 & 93.8 & 94.0 & 93.0 & 93.4 & 93.5 & 93.2 & 93.5 & 93.9 & 92.8 & 93.6 & 94.5 \\
\checkmark & \checkmark & \checkmark & \checkmark & \textbf{96.7} & \textbf{96.8} & \textbf{97.3} & \textbf{96.2} & \textbf{96.5} & \textbf{97.5} & \textbf{96.1} & \textbf{96.3} & \textbf{96.8} & \textbf{94.5} & \textbf{95.8} & \textbf{97.0}\\
\bottomrule
\end{tabular}
}
\label{table:loss}
\end{table*}

\subsection{Implementation Details for MedTrim}
MedTrim employed transformer-based encoders for projecting multimodal CXR data onto the latent embedding space. A ViT module pre-trained on natural images was taken as the image encoder \citep{dosovitskiy2020image}, and a BERT module pre-trained on natural text was taken as the text encoder \citep{2019arXiv190403323A}. 1,250,000 unique triplets were mined from the training set. For consistency, image samples in each triplet were curated from a single CXR view (anteroposterior, posteroanterior or lateral), and images were resized to 224$\times$224 pixels. Radiology reports that contained a findings or impressions section were curated. Hyperparameters for triplet mining were selected as $\gamma_0 = 0.85$, $\gamma_1 = 0.1$, $\gamma_2 = 0.05$, $\tau_{min}=0.25$, $\tau_{max}=0.6$, $\alpha=0.3$, and $\eta=0.5$ via cross-validation.

\subsection{Competing Methods}
MedTrim was comparatively demonstrated against state-of-the-art alignment methods for med-VLMs. For all methods, training was performed on NVidia RTX 4090 GPUs via the Adam optimizer, using a learning rate of $4 \times 10^{-5}$, a mini-batch size of 64, a weight decay factor of $1 \times 10^{-4}$, and 20 epochs.

\textbf{ConVIRT} \citep{convirt}: uses contrastive learning to align medical images and random sentences within radiology reports at the global level.

\textbf{JoImTeRNet} \citep{second_met}: uses contrastive learning for global image-sentence and local region-phrase alignment. Cross-entropy, ranking, masked modeling losses are combined.  

\textbf{GLoRIA} \citep{9710099}: uses contrastive learning for global image-text and local region-word alignment. Local and global contrastive losses are combined. 

\textbf{CheXzero} \citep{chexzero}: uses contrastive learning to align medical images and impressions in radiology reports.

\textbf{LIMITR} \citep{limitr}: uses knowledge-guided contrastive learning for image-report alignment at local and global levels. Location information and visual cues from multiple views (frontal and lateral) are employed for guidance.

\textbf{MedFILIP} \citep{medfilip}: uses knowledge-guided contrastive learning where an LLM extracts structured labels on disease class, severity and location from radiology reports. Image and text embeddings are aligned against a reference semantic similarity matrix derived from these labels.

\subsection{Performance Evaluation}
Models were first demonstrated for within-modal (image-to-image, text-to-text) and cross-modal (image-to-text, text-to-image) retrieval tasks. Separate evaluations were conducted to assess the consistency between query and retrieved items in terms of disease class, adjectival descriptor and directional descriptor. In each case, performance was measured via the Precision@R (P@R) metric with $R \in \{1, 10, 20, 50\}$. P@R reflects the percentage of consistent items among the top-$R$ retrieved items. Retrieved items (i.e., image or text) were ranked according to the cosine similarity between their embedding vectors. The level of consistency between the meta-entities of query and retrieved items was measured via Jaccard index. Models were also demonstrated for disease classification tasks from CXR images. Performance was measured via accuracy (ACC), area under the curve (AUC), and F1 metrics. For a given CXR image, the image embedding was compared against the embeddings of candidate text prompts in the form of “This is an X-Ray image of {disease}.”.  Disease class was predicted based on the prompt with the maximum similarity between image and text embeddings. For robust evaluations, the test set was split into 20 non-overlapping subsets in MIMIC-CXR and 5 non-overlapping subsets in CheXpert and RSNA Pneumonia. Metrics were then reported as mean$\pm$std across subsets. Statistical significance of differences in model performance were assessed via non-parametric Wilcoxon signed-rank tests.

\begin{figure}[t]
    \centering
    \includegraphics[width=0.75\linewidth]{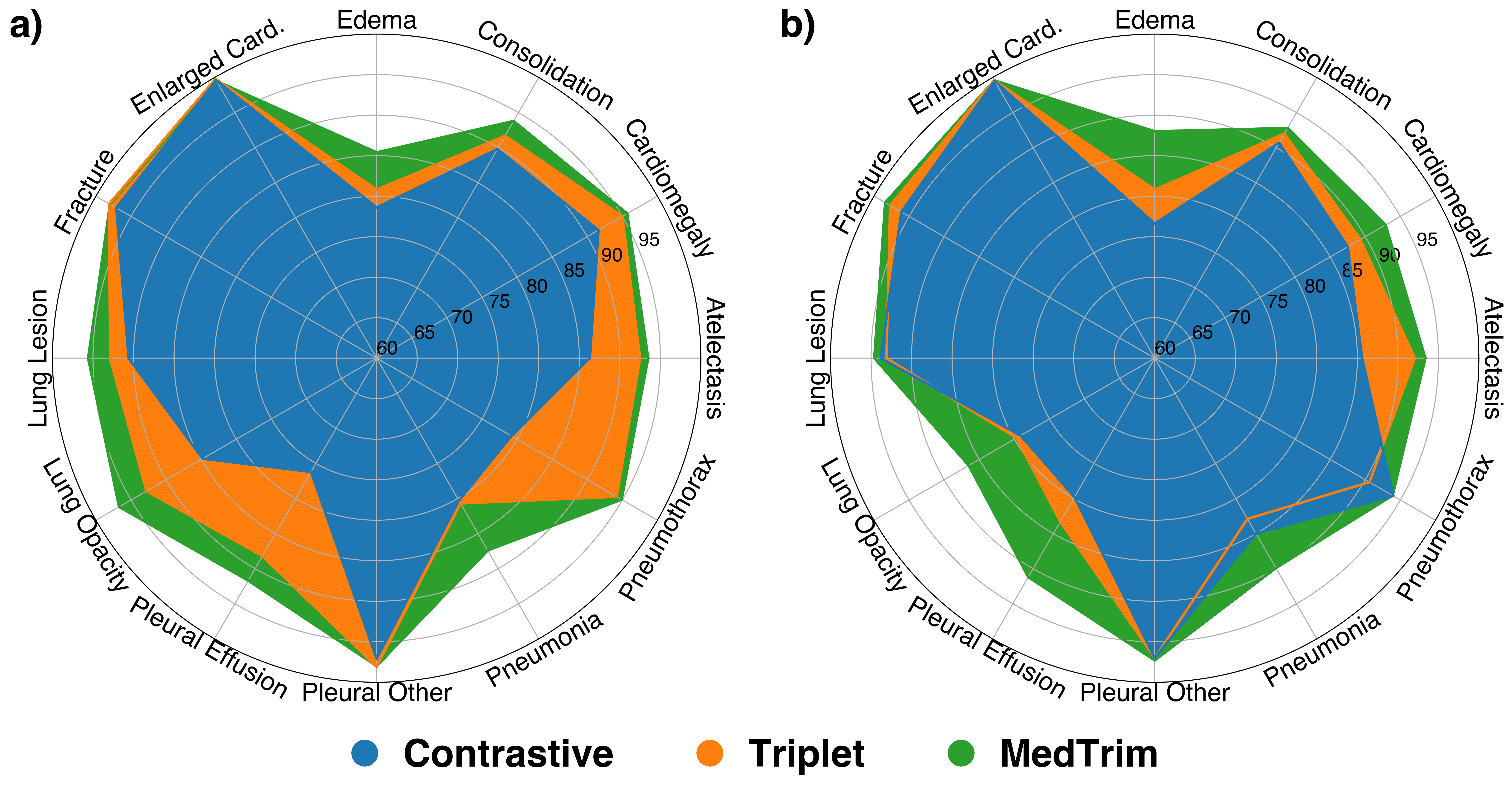}
    \caption{\textbf{Distribution of retrieval performance across disease classes.} Results shown for \textbf{(a)} I2T, \textbf{(b)} T2I retrieval tasks for disease class. MedTrim was compared against Contrastive and Triplet variants (see legend).}
    \label{fig:radar}
\end{figure}

\begin{figure}[t]
    \centering
    \includegraphics[width=\columnwidth]{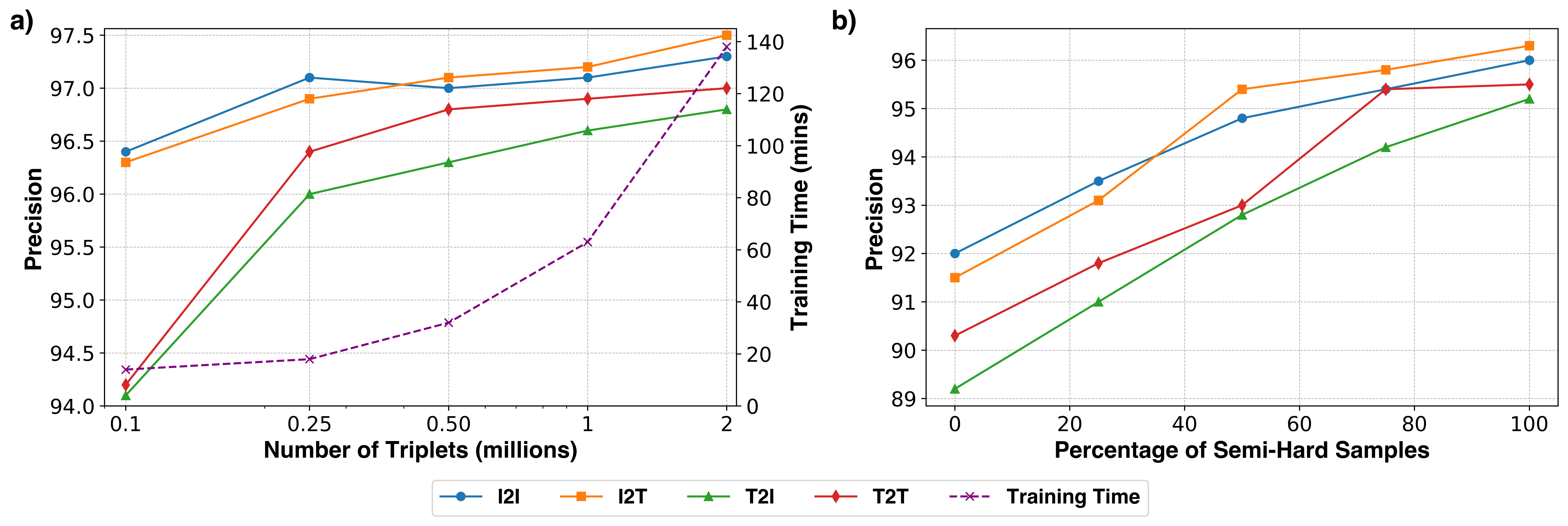}
    \caption{\textbf{(a) Ablation study on the number of mined triplets.} P@50 (\%; left y-axis) in I2I, I2T, T2I, and T2T tasks are plotted as a function of the number of triplets, along with the training time (min; right y-axis). \textbf{(b) Ablation study on the percentage of semi-hard negative samples.} P@50 in I2I, I2T, T2I, and T2T tasks are plotted as a function of the percentage of semi-hard negative samples included in triplets.}
    \label{fig:triplet_semi}
\end{figure}

\begin{table}[t]
\centering
\raisebox{-8mm}{%
  \begin{minipage}[t]{0.49\textwidth}
  \caption{Occurrence of unique disease classes in the MIMIC-CXR dataset expressed as percentage of all disease labels.}
  \centering
  \resizebox{\textwidth}{!}{%
  \begin{tabular}{l c l c}
      \toprule
      \textbf{Disease} & \textbf{Per. (\%)} & \textbf{Disease} & \textbf{Per. (\%)} \\
      \midrule
      Pleural Effusion & 19.0 & Consolidation & 3.6  \\
      Atelectasis & 18.2 & Pneumothorax & 2.5 \\
      Cardiomegaly & 16.7 & Lung Lesion & 2.1  \\
      Lung Opacity & 15.7 & Pleural Other & 1.3  \\
      Edema & 11.2 & Fracture & 1.0  \\
      Pneumonia & 8.5 & Enlarged Card. & 0.2 \\
      \bottomrule
  \end{tabular}
  }
  \label{tab:disease_freq}
  \end{minipage}
}
\hspace{2mm}  
\begin{minipage}[t]{0.40\textwidth}
\caption{\textbf{Ablation study on the similarity score range for semi-hard negative samples.} Average retrieval performances across meta-entities and retrieval tasks are listed for different configurations of \(\tau_{min}\) and \(\tau_{max}\). Bounds used in MedTrim are marked in gray.}
\centering
\resizebox{\textwidth}{!}{%
\begin{tabular}{ccccc}
\toprule
\(\boldsymbol{\tau_{min}}\) & \(\boldsymbol{\tau_{max}}\) & \textbf{P@10} & \textbf{P@20} & \textbf{P@50} \\
\midrule
0.10  & 0.60  & 95.0 & 95.5 & 96.1 \\
0.10  & 0.80  & 94.7 & 94.9 & 95.6 \\
\rowcolor[gray]{0.9} \textbf{0.25} & \textbf{0.60}  & \textbf{95.9} & \textbf{96.4} & \textbf{97.2} \\
0.25 & 0.80  & 95.8 & 96.3 & 96.8 \\
0.40  & 0.60  & 94.9 & 95.3 & 95.6 \\
0.40  & 0.80  & 94.6 & 95.0 & 95.2 \\
\bottomrule
\end{tabular}
}
\label{tab:tau_comparison}
\end{minipage}
\end{table}

\begin{table*}[t]
\centering
\caption{\textbf{Performance of competing methods in multimodal retrieval tasks.} Consistency between query and retrieved items was evaluated in terms of disease-class (Dis.), adjectival-descriptor (Adj.), and directional-descriptor (Dir.) entities. Precision values (P@R) for varying number of retrieved items (R) are listed as mean$\pm$std across test subsets in MIMIC-CXR.} 
\resizebox{1.0\columnwidth}{!}{%
\begin{tabular}{l l c c c c c c c c c c c c}
\toprule
 &  & \multicolumn{3}{c}{\textbf{P@1}} & \multicolumn{3}{c}{\textbf{P@10}} & \multicolumn{3}{c}{\textbf{P@20}} & \multicolumn{3}{c}{\textbf{P@50}} \\
\cmidrule(lr){3-5} \cmidrule(lr){6-8} \cmidrule(lr){9-11} \cmidrule(lr){12-14}
 & & \textbf{Disease} & \textbf{Adjective} & \textbf{Direction} & \textbf{Disease} & \textbf{Adjective} & \textbf{Direction} & \textbf{Disease} & \textbf{Adjective} & \textbf{Direction} & \textbf{Disease} & \textbf{Adjective} & \textbf{Direction} \\
\midrule
\multirow{7}{*}{\rotatebox[origin=c]{90}{\parbox[c]{1cm}{\centering \textbf{I2I}}}}
  & ConVIRT       & 90.3$\pm$2.4 & 77.8$\pm$4.4 & 80.2$\pm$3.9 & 90.6$\pm$2.3 & 78.1$\pm$4.2 & 80.6$\pm$3.5 & 90.8$\pm$2.3 & 78.0$\pm$4.0 & 81.0$\pm$3.5 & 90.8$\pm$2.4 & 78.5$\pm$4.1 & 81.2$\pm$3.3 \\
  & JoImTeRNet    & 90.0$\pm$2.6 & 65.3$\pm$6.5 & 72.4$\pm$5.8 & 90.5$\pm$2.7 & 65.3$\pm$6.3 & 72.0$\pm$5.6 & 89.3$\pm$2.7 & 66.3$\pm$6.2 & 73.0$\pm$5.7 & 89.0$\pm$2.7 & 66.8$\pm$5.9 & 73.5$\pm$5.6 \\
  & GLoRIA        & 90.0$\pm$2.6 & 75.0$\pm$5.1 & 80.0$\pm$4.1 & 89.0$\pm$2.8 & 72.5$\pm$4.8 & 78.0$\pm$4.5 & 88.0$\pm$2.6 & 74.0$\pm$4.5 & 80.0$\pm$3.9 & 90.0$\pm$2.3 & 80.0$\pm$3.8 & 80.0$\pm$3.7 \\
  & CheXzero      & 92.0$\pm$2.4 & 77.0$\pm$4.7 & 82.0$\pm$3.8 & 92.5$\pm$2.4 & 77.5$\pm$4.5 & 82.5$\pm$3.6 & 92.3$\pm$2.6 & 77.3$\pm$4.9 & 82.3$\pm$3.6 & 92.2$\pm$2.5 & 77.2$\pm$4.5 & 82.2$\pm$3.5 \\
  & LIMITR        & 93.0$\pm$2.3 & 80.4$\pm$4.0 & 83.1$\pm$3.5 & 93.2$\pm$2.3 & 81.0$\pm$3.5 & 83.2$\pm$3.6 & 94.0$\pm$2.1 & 82.6$\pm$3.3 & 84.9$\pm$3.3 & 94.2$\pm$2.0 & 83.0$\pm$3.4 & 85.0$\pm$3.3 \\
  & MedFILIP      & 95.5$\pm$1.8 & 82.5$\pm$3.5 & 87.8$\pm$3.3 & 96.1$\pm$1.8 & 84.2$\pm$3.3 & 88.5$\pm$3.0 & 96.4$\pm$1.6 & 84.7$\pm$3.0 & 89.0$\pm$2.8 & 96.9$\pm$1.5 & 86.0$\pm$2.4 & 90.8$\pm$2.4 \\
  \rowcolor[gray]{0.9} & MedTrim & \textbf{96.4$\pm$1.8} & \textbf{83.9$\pm$4.1} & \textbf{90.0$\pm$2.3} & \textbf{96.7$\pm$1.6} & \textbf{85.1$\pm$3.5} & \textbf{91.0$\pm$2.1} & \textbf{96.8$\pm$1.6} & \textbf{85.6$\pm$3.1} & \textbf{91.2$\pm$2.0} & \textbf{97.3$\pm$1.4} & \textbf{87.7$\pm$2.8} & \textbf{92.6$\pm$2.0} \\
\midrule
\multirow{7}{*}{\rotatebox[origin=c]{90}{\parbox[c]{1cm}{\centering \textbf{I2T}}}}
  & ConVIRT       & 89.8$\pm$2.7 & 76.8$\pm$4.4 & 78.8$\pm$3.6 & 90.2$\pm$2.5 & 77.5$\pm$4.1 & 80.0$\pm$3.8 & 90.0$\pm$2.6 & 78.0$\pm$3.9 & 80.5$\pm$3.9 & 89.8$\pm$2.7 & 78.2$\pm$4.0 & 80.3$\pm$3.7 \\
  & JoImTeRNet    & 90.1$\pm$2.4 & 65.3$\pm$6.5 & 71.8$\pm$5.4 & 90.0$\pm$2.7 & 65.3$\pm$5.9 & 72.7$\pm$4.9 & 89.5$\pm$2.8 & 66.5$\pm$6.1 & 73.0$\pm$4.6 & 89.0$\pm$2.5 & 66.2$\pm$5.8 & 72.9$\pm$4.7 \\
  & GLoRIA        & 90.5$\pm$2.4 & 76.0$\pm$4.8 & 81.0$\pm$3.9 & 90.0$\pm$2.4 & 73.0$\pm$4.6 & 80.0$\pm$4.1 & 90.3$\pm$2.3 & 76.0$\pm$4.9 & 83.0$\pm$3.9 & 90.5$\pm$2.5 & 76.0$\pm$4.8 & 83.0$\pm$3.7 \\
  & CheXzero      & 93.0$\pm$2.3 & 79.0$\pm$4.4 & 84.0$\pm$3.6 & 93.5$\pm$2.6 & 79.5$\pm$4.1 & 84.5$\pm$3.2 & 93.3$\pm$2.3 & 79.3$\pm$4.5 & 84.3$\pm$3.5 & 93.2$\pm$2.2 & 79.2$\pm$4.3 & 84.2$\pm$3.3 \\
  & LIMITR        & 94.0$\pm$2.3 & 82.0$\pm$3.5 & 85.6$\pm$3.5 & 94.3$\pm$2.3 & 82.4$\pm$3.6 & 85.5$\pm$3.4 & 95.0$\pm$2.1 & 84.0$\pm$3.4 & 87.0$\pm$3.3 & 95.3$\pm$1.9 & 84.1$\pm$3.4 & 85.0$\pm$3.4 \\
  & MedFILIP      & 92.6$\pm$2.5 & 73.5$\pm$4.8 & 82.5$\pm$3.6 & 95.2$\pm$2.5 & 79.8$\pm$4.7 & 86.1$\pm$3.4 & 95.4$\pm$2.4 & 80.3$\pm$4.4 & 86.6$\pm$3.0 & 95.8$\pm$2.2 & 81.0$\pm$3.8 & 87.4$\pm$2.7 \\
  \rowcolor[gray]{0.9} & MedTrim & \textbf{96.2$\pm$1.8} & \textbf{82.9$\pm$3.2} & \textbf{89.2$\pm$2.1} & \textbf{96.2$\pm$1.9} & \textbf{85.5$\pm$2.7} & \textbf{89.5$\pm$2.2} & \textbf{96.5$\pm$1.8} & \textbf{88.3$\pm$2.3} & \textbf{90.2$\pm$2.0} & \textbf{97.5$\pm$1.6} & \textbf{90.7$\pm$2.0} & \textbf{93.3$\pm$1.9} \\
\midrule
\multirow{7}{*}{\rotatebox[origin=c]{90}{\parbox[c]{1cm}{\centering \textbf{T2I}}}}
  & ConVIRT       & 91.6$\pm$2.6 & 81.3$\pm$3.3 & 81.0$\pm$3.4 & 92.0$\pm$2.6 & 81.5$\pm$3.3 & 81.7$\pm$3.5 & 91.7$\pm$2.8 & 82.0$\pm$3.0 & 82.3$\pm$3.1 & 92.2$\pm$2.6 & 82.3$\pm$3.2 & 83.5$\pm$3.3 \\
  & JoImTeRNet    & 88.3$\pm$2.9 & 63.0$\pm$5.5 & 75.0$\pm$5.1 & 88.0$\pm$2.6 & 63.5$\pm$5.2 & 76.0$\pm$4.8 & 87.4$\pm$2.7 & 62.0$\pm$5.3 & 74.0$\pm$4.7 & 87.6$\pm$2.7 & 63.0$\pm$5.2 & 77.2$\pm$4.6 \\
  & GLoRIA        & 92.0$\pm$2.8 & 78.0$\pm$4.7 & 85.0$\pm$4.1 & 92.0$\pm$2.8 & 78.5$\pm$4.8 & 85.0$\pm$4.0 & 92.3$\pm$2.6 & 78.0$\pm$4.9 & 85.0$\pm$3.8 & 92.0$\pm$2.6 & 78.0$\pm$4.6 & 85.0$\pm$3.5 \\
  & CheXzero      & 93.5$\pm$2.3 & 80.0$\pm$4.0 & 86.0$\pm$3.2 & 93.5$\pm$2.1 & 80.5$\pm$3.8 & 86.0$\pm$3.0 & 93.8$\pm$2.1 & 80.3$\pm$3.5 & 86.0$\pm$3.1 & 93.5$\pm$1.9 & 80.2$\pm$3.4 & 86.0$\pm$3.0 \\
  & LIMITR        & 94.0$\pm$2.1 & 80.0$\pm$3.4 & 86.1$\pm$3.1 & 94.3$\pm$2.2 & 82.0$\pm$3.0 & 85.5$\pm$3.0 & 94.9$\pm$1.9 & 83.3$\pm$2.8 & 87.0$\pm$2.5 & 95.2$\pm$1.9 & 84.8$\pm$2.6 & 88.5$\pm$2.4 \\
  & MedFILIP      & 92.8$\pm$2.4 & 69.0$\pm$4.7 & 79.8$\pm$3.8 & 93.1$\pm$2.3 & 70.3$\pm$4.5 & 81.0$\pm$3.6 & 93.2$\pm$2.5 & 72.3$\pm$4.3 & 82.0$\pm$3.2 & 93.6$\pm$2.3 & 73.4$\pm$4.0 & 84.2$\pm$3.2 \\
  \rowcolor[gray]{0.9} & MedTrim & \textbf{95.9$\pm$2.0} & \textbf{80.3$\pm$3.0} & \textbf{87.3$\pm$2.4} & \textbf{96.1$\pm$1.6} & \textbf{83.8$\pm$2.6} & \textbf{89.4$\pm$2.1} & \textbf{96.3$\pm$1.6} & \textbf{85.4$\pm$2.4} & \textbf{90.0$\pm$2.4} & \textbf{96.8$\pm$1.4} & \textbf{86.0$\pm$2.5} & \textbf{91.7$\pm$2.1} \\
\midrule
\multirow{7}{*}{\rotatebox[origin=c]{90}{\parbox[c]{1cm}{\centering \textbf{T2T}}}}
  & ConVIRT       & 92.2$\pm$2.4 & 80.3$\pm$3.1 & 80.5$\pm$3.2 & 92.8$\pm$2.3 & 80.9$\pm$3.0 & 82.2$\pm$3.0 & 93.0$\pm$2.3 & 81.4$\pm$3.2 & 83.1$\pm$3.1 & 93.0$\pm$2.2 & 82.0$\pm$3.0 & 83.5$\pm$3.1 \\
  & JoImTeRNet    & 91.0$\pm$2.5 & 64.3$\pm$5.3 & 73.0$\pm$4.8 & 91.2$\pm$2.8 & 63.0$\pm$5.1 & 74.6$\pm$4.6 & 91.0$\pm$2.6 & 63.5$\pm$5.4 & 74.0$\pm$4.8 & 90.7$\pm$2.5 & 63.0$\pm$5.3 & 73.8$\pm$4.6 \\
  & GLoRIA        & 92.0$\pm$2.5 & 78.0$\pm$4.8 & 84.0$\pm$3.6 & 92.0$\pm$2.7 & 78.5$\pm$4.8 & 84.5$\pm$3.5 & 92.3$\pm$2.3 & 78.3$\pm$4.6 & 84.7$\pm$3.1 & 92.0$\pm$3.0 & 78.2$\pm$4.5 & 84.9$\pm$2.9 \\
  & CheXzero      & 93.0$\pm$2.2 & 79.0$\pm$4.3 & 85.0$\pm$3.6 & 93.0$\pm$2.5 & 79.5$\pm$4.1 & 85.0$\pm$2.3 & 93.5$\pm$1.9 & 79.3$\pm$4.1 & 85.0$\pm$2.2 & 93.0$\pm$2.0 & 79.2$\pm$4.1 & 85.0$\pm$2.4 \\
  & LIMITR        & 93.7$\pm$2.4 & 81.5$\pm$3.2 & \textbf{85.9$\pm$3.3} & 94.1$\pm$2.8 & 82.0$\pm$3.8 & 87.0$\pm$2.4 & 94.4$\pm$2.0 & 83.0$\pm$3.6 & 87.6$\pm$3.4 & 95.1$\pm$1.6 & 84.0$\pm$3.1 & 88.0$\pm$3.3 \\
  & MedFILIP      & 94.0$\pm$1.6 & 82.4$\pm$3.2 & 86.2$\pm$2.9 & \textbf{95.2$\pm$1.6} & 83.4$\pm$3.2 & 87.0$\pm$2.9 & 95.6$\pm$2.0 & 84.3$\pm$2.8 & 87.9$\pm$2.6 & 96.5$\pm$1.8 & 85.6$\pm$2.3 & 90.4$\pm$3.0 \\
  \rowcolor[gray]{0.9} & MedTrim & \textbf{94.3$\pm$1.8} & \textbf{83.1$\pm$3.0} & 85.6$\pm$2.0 & 94.5$\pm$1.4 & \textbf{83.6$\pm$2.1} & \textbf{87.4$\pm$2.1} & \textbf{95.8$\pm$1.3} & \textbf{84.7$\pm$3.1} & \textbf{88.8$\pm$2.2} & \textbf{97.0$\pm$1.5} & \textbf{86.7$\pm$3.1} & \textbf{91.1$\pm$2.7} \\
\bottomrule
\end{tabular}
}
\label{table:comparison_retrieval}
\end{table*}

\begin{figure*}[t]
\vspace{-0.2cm}
    \centering
    \includegraphics[width=0.75\textwidth]{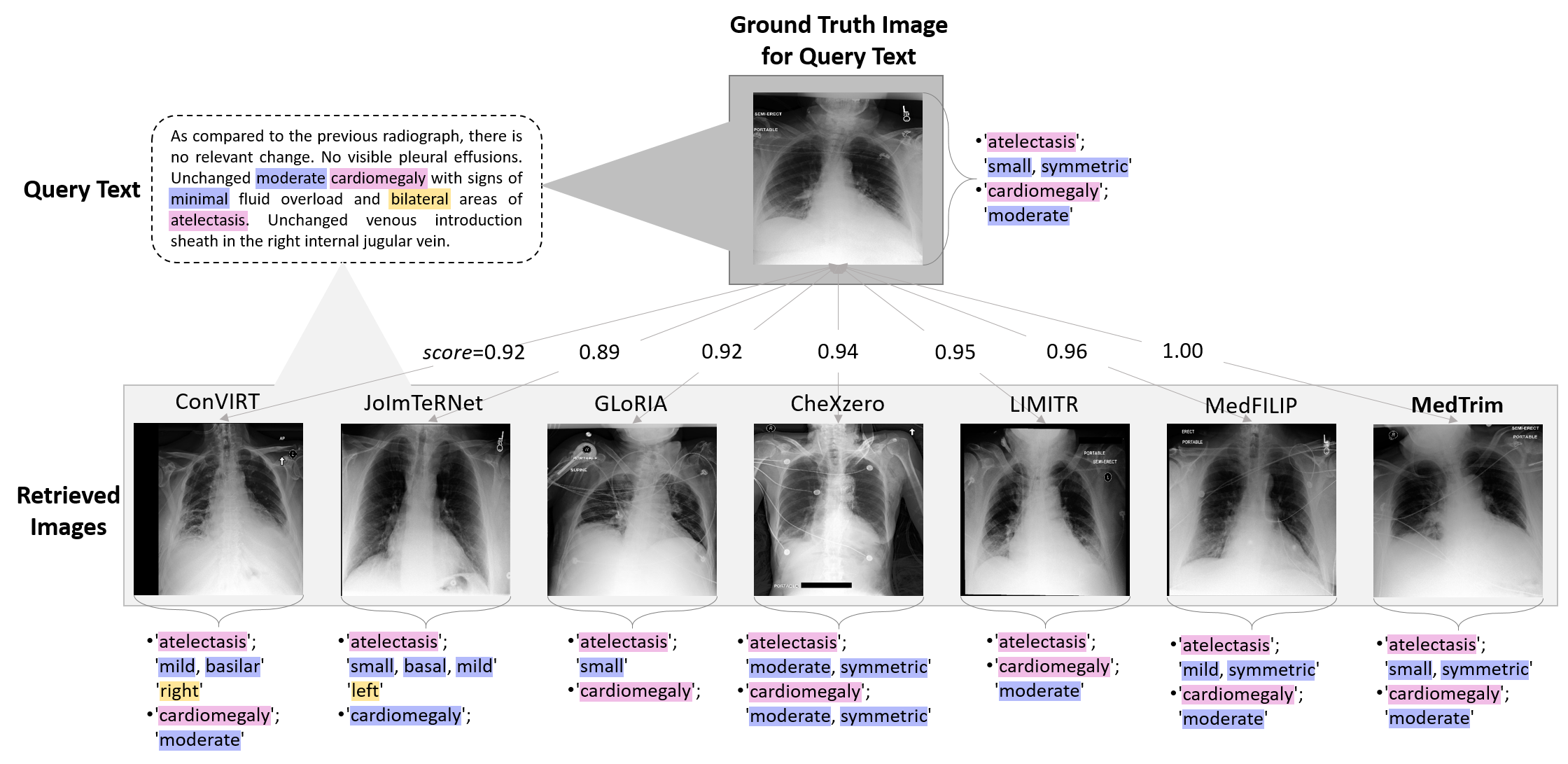}
    \vspace{-0.2cm}
    \caption{\textbf{Representative results for T2I retrieval.} Given a query CXR report, the top-ranked images retrieved by competing methods are displayed, along with their similarity scores to the ground-truth CXR image. For each method, meta-entities extracted from the reports of the retrieved images are highlighted (magenta: disease class, purple: adjectival descriptor, yellow: directional descriptor).}
    \label{fig:visual_retrieval}
\end{figure*}
\begin{figure*}[th!]
    \vspace{-0.2cm}
    \centering
    \includegraphics[width=\textwidth]{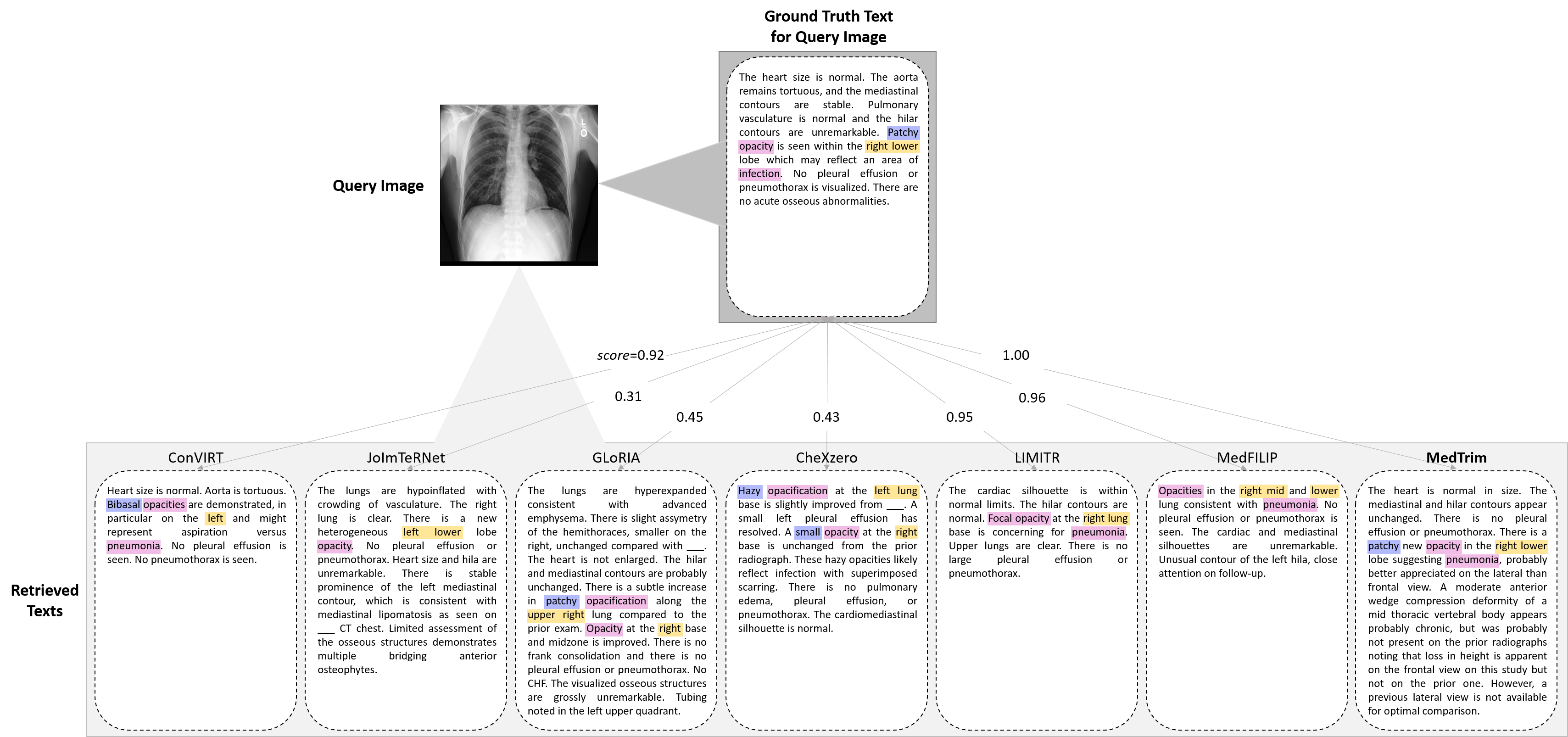}
        \vspace{-0.4cm}
    \caption{\textbf{Representative results for I2T retrieval.} Given a query CXR image, the top-ranked reports retrieved by competing methods are displayed, along with their similarity scores to the ground-truth CXR report. For each method, meta-entities in the retrieved reports are highlighted (magenta: disease class, purple: adjectival descriptor, yellow: directional descriptor).}
    \label{fig:visual_retrieval2}
\end{figure*}

\section{Results}
\subsection{Ablation Studies}

\textbf{Meta-entity guidance:} To assess the impact of meta-entity guidance on the performance of MedTrim, we conducted an ablation study by training variant models that selectively enabled or disabled guidance from disease classes, adjectival descriptors, and directional descriptors. Models were evaluated for downstream image-to-image (I2I), image-to-text (I2T), text-to-image (T2I) and text-to-text (T2T) retrieval tasks. Table~\ref{table:metaentity} lists average retrieval precision (P@R with R denoting the number of retrieved items) of variant models across tasks, while separately evaluating consistency between query and retrieved items in terms of disease-class, adjectival-descriptor and directional-descriptor entities. On average, removing adjectival guidance leads to a 1.7\% drop in precision for adjectival descriptors, removing directional guidance leads to a 1.6\% drop in precision for directional descriptors, and removing both elicits a 2.5\% drop for adjectival and 2.6\% drop for directional descriptors. These results validate our hypothesis that incorporating structured meta-entity guidance into triplet learning elicits attribute-aware alignment of image-text representations, which in turn improves retrieval performance in multimodal tasks.

\vspace{0.8ex} \textbf{Multimodal triplet objective:} To assess the impact of the multimodal triplet alignment objective, we trained variant models that selectively ablated different objective components, including the base triplet-type loss ($f_{tri}$), within-modal loss terms ($\mathcal{L}_{I2I+T2T}$) and cross-modal loss terms ($\mathcal{L}_{I2T},\mathcal{L}_{T2I}$). Variants ablating triplet-type loss replaced $f_{tri}$ with a contrastive-type loss between anchor-positive and anchor-negative pairs. Models were evaluated for multimodal retrieval tasks, and precision was measured separately for disease class, and adjectival and directional descriptors. Table~\ref{table:loss} lists average retrieval precision across meta-entities, separately for I2I, I2T, T2I, T2T tasks. In general, triplet-type variants outperform contrastive-type variants when the two types are implemented with matching sets of within- and cross-modal loss terms. On average, triplet variants offer 6.2\% higher precision than contrastive variants. Among triplet variants, inclusion of either $\mathcal{L}_{I2T}$ or $\mathcal{L}_{T2I}$ enhances precision over the variant that only uses within-modal terms, with an average 8.1\% improvement. MedTrim using all loss terms achieves the highest scores, with 10.8\% improvement over the variant containing only within-modal terms and 2.9\% improvement over the variant containing only cross-modal terms. These findings confirm that the proposed triplet learning framework enables enhanced image-text alignment for CXR data compared to contrastive learning, and that within- and cross-modal alignment objectives serve complementary roles in refining representational learning in med-VLMs.

\textbf{Class-distribution of retrieval performance:} A fundamental challenge in medical image-text retrieval is addressing class imbalance, where certain diseases are disproportionately represented compared to others. Notably, atelectasis and pleural effusion are among the most frequently occurring conditions, whereas others, such as enlarged cardiomediastinum and fracture, can be relatively rare. In learning-based retrieval, this imbalance can bias retrieved items towards either very common or rare disease classes, compromising precision for retrieval of remaining classes. Since consistent precision across disease classes is a crucial aspect of effective retrieval systems, we examined the distribution of retrieval performance across individual disease classes. Fig. \ref{fig:radar} depicts MedTrim's retrieval precision across disease classes for I2T and T2I tasks, against a contrastive variant that used the same objective albeit ablated $f_{tri}$, and a triplet variant that only used cross-modal loss terms. While all models show moderate variations, MedTrim attains consistently high precision across disease classes, and superior precision against variant models in moderately frequent disease classes such as edema, pneumonia, consolidation or lung lesion. These results highlight the importance of MedTrim's multimodal triplet objective in achieving more equitable retrieval performance under class imbalances.

\textbf{Extent of triplet mining:} The number of triplets used to optimize the alignment objective of MedTrim can play a crucial role in shaping the quality of the learned embedding space. While training on an excessive number of triplets could theoretically lead to an ideal representation, such an approach is computationally infeasible. To balance performance and efficiency, we conducted experiments where MedTrim was trained using 0.1 to 2 million unique triplets, derived from the training set. Fig.~\ref{fig:triplet_semi}a displays precision in multimodal retrieval tasks for varying numbers of triplets. Following an initial inclination towards 0.25 million triplets, diminishing returns are observed in performance for further increases in triplet count, consistently across different retrieval tasks. Note that training time elevates from just under 20 minutes for 0.25 million triplets to over 2 hours for 2 million triplets. This substantial increase in computational cost, coupled with marginal performance gains, suggests that approximately 0.25 million triplets strikes an optimal balance between retrieval precision and training efficiency.

\textbf{Semi-hard negative sampling:} A critical factor that influences the ability of triplet mining to learn discriminative representations is the selection of negative samples. To improve learning of intra-class variations, MedTrim employs semi-hard negative samples by constraining the similarity scores between anchor and negative samples to a moderate range. This constraint serves to ensure that selected negatives share some overlap with the anchor in terms of disease classes (e.g., at least one shared disease) while differing in adjectival or directional descriptors. We conducted two ablation studies to examine the importance of semi-hard negative sampling in MedTrim. First, we built variant models trained on a mixture of semi-hard negative samples and easy negative samples. Fig. \ref{fig:triplet_semi}b displays precision in multimodal retrieval tasks for varying percentages of semi-hard samples in the training set. We find that increasing the proportion of semi-hard negatives significantly improves performance in both within-modal and cross-modal retrieval tasks. Second, we build variant models trained using negative samples with similarity scores constrained to different sets of $\tau_{min,\,max}$. Table~\ref{tab:tau_comparison} lists average retrieval precision for variants. We observe that the proposed score range attains optimal performance among the examined ranges. Collectively, these results underscore the importance of using challenging but informative negatives for enhanced triplet learning.

\begin{figure*}[h]
\vspace{-0.2cm}
    \centering
    \includegraphics[width=\textwidth]{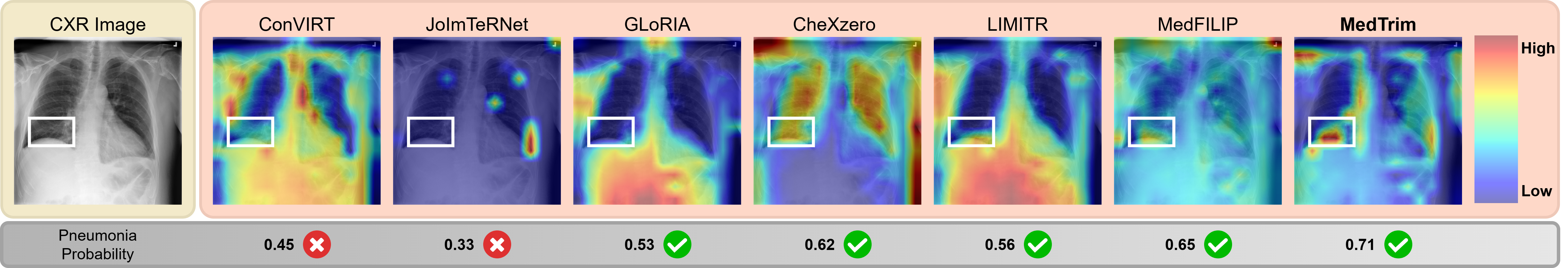}
    \vspace{-0.2cm}
    \caption{\textbf{Representative results for zero-shot disease classification.} For an input CXR image, Grad-CAM heatmaps (see colorbar) highlighting regions with high contribution to the classification decision are displayed for competing methods, along with confidence scores (i.e., probability of pneumonia). White boxes denote the radiologist-annotated region-of-interest containing pathology in the CXR image.}
    \label{fig:visual_classification}
\end{figure*}

\subsection{Comparison Studies}

\subsubsection{Multimodal Retrieval Tasks}
Next, we demonstrated MedTrim via comparisons against state-of-the-art alignment methods for med-VLMs. In particular, MedTrim was compared against global baselines (ConVIRT, and CheXzero), global-local baselines (JoImTeRNet, and GLoRIA), and knowledge-guided baselines (LIMITR, and MedFILIP). Methods were evaluated on the MIMIC-CXR dataset to perform multimodal retrieval tasks. Precision values are listed in Table \ref{table:comparison_retrieval}, separately evaluating consistency of query and retrieved items in terms of disease class, adjectival descriptor, and directional descriptor entities. MedTrim achieves the highest retrieval performance in all examined settings (p$<$0.05), except for LIMITR that yields higher directional P@1, and MedFILIP that yields higher disease P@10 in T2T. On average, MedTrim offers improvements of 5.7\% precision over global baselines, 10.7\% precision over global/local baselines, and 3.3\% precision over knowledge-guided baselines. Representative retrieval results from competing methods are depicted in Fig. \ref{fig:visual_retrieval} for T2I and in Fig. \ref{fig:visual_retrieval2} for I2T tasks. Compared to baselines, we observe that MedTrim retrieves items with higher similarity scores to the sought ground-truth information. In T2I, MedTrim retrieves an image whose report contains more closely matched meta-entities to the query report. In I2T, MedTrim retrieves a report whose meta-entities are more closely matched to the ground-truth report of the query image. These results demonstrate the superior ability of MedTrim to align and retrieve diagnostically-relevant cross-modal information in CXR data.

\subsubsection{Disease Classification Tasks} We also demonstrated MedTrim via comparisons against state-of-the-art alignment methods for zero-shot disease classification tasks on the CheXpert and RSNA Pneumonia datasets. Accuracy, AUC and F1 values for competing methods are listed in Table \ref{tab:comparison_classification}. MedTrim achieves the highest performance metrics among competing methods (p$<$0.05). On average, MedTrim offers (ACC, AUC, F1) improvements of (5.9\%, 6.3\%, 6.0\%) over global baselines, (9.7\%, 13.8\%, 8.8\%) over global/local baselines, and (3.0\%, 3.1\%, 4.6\%) over knowledge-guided baselines. Representative classification results from competing methods, with Grad-CAM activation maps overlaid onto CXR images \citep{gradcam}, are depicted in Fig. \ref{fig:visual_classification}. We observe that the activation maps for MedTrim more closely focus on diagnostically-relevant regions of the CXR image containing pathology, whereas the activation maps are more diffusely distributed across the field-of-view for baselines. Overall, these results demonstrate the superior ability of MedTrim to detect disease-related features in CXR images, without requiring dataset-specific supervision.

\section{Discussion}
Our demonstrations on MedTrim highlight the potential of triplet learning for multimodal image-text alignment in med-VLMs, but several technical limitations remain, offering avenues for future work. A first avenue concerns the nature of multimodal data available for model training. The current implementation of MedTrim assumes access to a training dataset where CXR images and respective radiology reports are available for each subject. While such datasets exist publicly \citep{johnson2019mimic,irvin2019chexpert}, single-modality datasets containing only CXR images or only radiology reports are relatively more common \citep{doi:10.1148/ryai.2019180041}. To seek performance benefits by enabling training on broader datasets, triplet learning can be augmented with cycle-consistency losses \citep{10167641}. For this purpose, a translation module could be used that bidirectionally maps between images and text in the latent embedding space to permit capture of meaningful cross-modal representations.  

A key component of MedTrim is extraction of pathology attributes as meta-entities from CXR reports, which guides triplet mining by selecting informative positive and semi-hard negative samples. In the current implementation, these attributes are extracted from radiology reports and used to align entire images and their associated text. This can be seen as a hybrid approach, enforcing global alignment for images while considering a common ground between global and local alignment for text \citep{10376847}. Future work could explore enhanced localization by integrating region-identification or patch-based techniques on medical images \citep{second_met}, which have been shown to improve spatial precision in multimodal learning. The meta-entities in the current implementation were based on adjectival and directional descriptors, as they are among the most commonly used attributes in radiology reports for describing CXR findings. However, additional structured information—such as external medical ontologies or dictionaries—could be incorporated into the triplet mining process. Enrichment of the entity definitions with external knowledge sources might enable further precision improvements in multimodal alignment.

\begin{table}[t]
\centering
\setlength\tabcolsep{4pt}
\renewcommand{\arraystretch}{1.4}
\caption{\textbf{Performance of competing methods in disease classification tasks.} Zero-shot classification was conducted on CheXpert (5 classes, 20\% chance level) and RSNA (2 classes, with 20\% positive-class and 80\% negative-class chance levels) datasets. Accuracy (ACC), area-under-the-curve (AUC) and F1 metrics are listed.}
\resizebox{0.6\columnwidth}{!}{%
\begin{tabular}{l ccc ccc}
\toprule
 & \multicolumn{3}{c}{\textbf{CheXpert}} & \multicolumn{3}{c}{\textbf{RSNA}} \\
\cmidrule(lr){2-4} \cmidrule(lr){5-7}
       & \textbf{ACC} & \textbf{F1} & \textbf{AUC} 
       & \textbf{ACC} & \textbf{F1} & \textbf{AUC}  \\
\midrule
ConVIRT     & 36.4$\pm$2.5 & 36.7$\pm$2.4 & 64.1$\pm$1.4 & 71.9$\pm$1.6 & 68.6$\pm$1.2 & 55.8$\pm$3.1 \\ 
JoImTeRNet  & 28.9$\pm$1.5 & 29.4$\pm$2.6 & 59.8$\pm$2.5 & 69.4$\pm$1.8 & 66.3$\pm$1.4 & 51.4$\pm$3.4 \\ 
GLoRIA      & 38.6$\pm$2.2 & 39.4$\pm$2.4 & 66.1$\pm$1.4 & 74.1$\pm$1.3 & 71.2$\pm$1.2 & 59.2$\pm$2.8 \\ 
CheXzero    & 40.0$\pm$1.9 & 40.6$\pm$2.1 & 66.3$\pm$1.3 & 78.0$\pm$1.1 & 76.9$\pm$1.0 & 66.4$\pm$1.9 \\ 
LIMITR      & 42.0$\pm$1.7 & 42.5$\pm$2.1 & 67.2$\pm$1.2 & 76.3$\pm$1.4 & 74.2$\pm$1.3 & 61.9$\pm$2.2 \\ 
MedFILIP    & 41.8$\pm$2.1 & 42.4$\pm$2.2 & 66.7$\pm$1.1 & 77.8$\pm$1.5 & 76.6$\pm$1.3 & 62.3$\pm$1.8 \\ 
\rowcolor[gray]{0.9} 
MedTrim     & \textbf{45.1$\pm$1.7} & \textbf{46.9$\pm$2.1} & \textbf{70.0$\pm$1.1} & \textbf{79.8$\pm$1.2} & \textbf{77.1$\pm$0.9} & \textbf{68.3$\pm$1.9} \\ 
\bottomrule
\end{tabular}
}
\label{tab:comparison_classification}
\end{table}

In this study, we primarily focused on zero-shot deployment of aligned med-VLMs to downstream retrieval and classification tasks, where we observed that MedTrim shows superior generalization to different tasks and datasets than competing methods. Prior research suggests that generalization to different datasets, particularly when they contain previously unseen disease classes or attributes, can benefit from few-shot learning on a small set of samples from the target domain \citep{pmlr-v139-radford21a,2023arXiv230519894W,PARK2024103021,10182304}. To improve reliability of few-shot learning, semantic-similarity measures between disease-descriptive entities could be explicitly computed, and injected into the triplet learning objective to establish stronger links \citep{10.1007/978-3-031-67751-9_7,10376864,YANG202271}. Such additional guidance might improve MedTrim’s ability to handle rare or novel disease cases.

Beyond retrieval and classification, other downstream tasks could benefit from the improved multimodal alignment provided by MedTrim. One promising direction is to attach modality-specific decoders to reconstruct images or radiology reports from the learned embedding vectors in med-VLMs, allowing us to explore the proposed technique's potential for medical image or text generation \citep{9768661,info16020136,yin-etal-2025-kia,arslan2024selfconsistentrecursivediffusionbridge, kabas2024physicsdrivenautoregressivestatespace}. Alternatively, task-specific models could be trained on top of the aligned embeddings for applications such as segmentation, severity grading, or report summarization \citep{9638337}. Given that interpretation of multimodal data is a general component of all imaging-based clinical assessments, MedTrim's framework could be extended to other imaging modalities, such as MRI or CT, to allow broader applications in medical imaging.

\section{Conclusion}
Representational alignment between medical images and radiology reports is key to building performant medical vision-language models (med-VLMs). Here, we introduced a novel triplet learning framework, MedTrim, to enable explicit guidance from disease classes and pathology descriptors in refining sample selection, so as to derive image-text embeddings that capture fine-grained intra-class variability, while reliably segregating different classes. Our experiments indicate that this approach elicits superior retrieval and classification performance on CXR data over state-of-the-art alignment methods. Therefore, MedTrim holds great promise for training reliable med-VLMs with precise image-text alignment and thereby improved sensitivity in downstream tasks. 

\section*{Acknowledgment}
This work was supported in part by TUBA GEBIP 2015 and BAGEP 2017 fellowships awarded to T. \c{C}ukur, in part by a BAGEP 2023 fellowship awarded to A. Ko\c{c}, and in part by TUBITAK Grant no. 123E142 awarded to T. \c{C}ukur.

\bibliographystyle{unsrt}  
\bibliography{main}

\end{document}